\documentclass[aos,preprint]{imsart}
\usepackage{bookmark}

\usepackage{amsthm,amsmath,amsfonts,amssymb}
\usepackage{enumerate}
\usepackage{multirow}
\usepackage{longtable}
\usepackage{bm,accents}
\usepackage{graphicx}
\hypersetup{pdfstartview={XYZ null null 1}}
\usepackage{url}
\usepackage{breakurl}
\usepackage{color}

\usepackage[square]{natbib}
\usepackage{xr}
\allowdisplaybreaks

\startlocaldefs

\newtheorem{thm}{Theorem}
\newtheorem{cor}[thm]{Corollary}
\newtheorem{lem}[thm]{Lemma}

\newtheorem{prop}[thm]{Proposition}

\newtheorem{algo}{Algorithm}
\newtheorem{cond}{Condition}
\newtheorem{defn}{Definition}
\theoremstyle{remark}
\newtheorem{example}{Example}
\newtheorem{rem}{Remark}
\usepackage{chngcntr}
\counterwithin{table}{section}

\setattribute{tablename} {skip}{.~}
\catcode`~=11 
\newcommand{\urltilde}{\kern -.15em\lower .7ex\hbox{~}\kern .04em}
\catcode`~=13 

\begin{document}

\sloppy
\begin{frontmatter}


\title{Feature Elimination in Kernel Machines in moderately high dimensions}
\runtitle{A notion of consistency for RFE in kernel machines}
\thankstext{T1}{The authors are grateful to the anonymous reviewers, the associate editor, and the editor for their helpful suggestions and comments. This research was funded in part by Israel Science Foundation grant 1308/12 and partly by United States NCI grant CA142538. The first author is thankful to the Friday Machine Learning Lab at the Biostatistics Department, University of North Carolina at Chapel Hill, for many helpful discussions and suggestions.}

\begin{aug}
\author[*]{\fnms{Sayan} \snm{Dasgupta}\ead[label=e1]{sdg.roopkund@gmail.com}},
\author[**]{\fnms{Yair} \snm{Goldberg}\ead[label=e2]{ygoldberg@stat.haifa.ac.il}} and
\author[*]{\fnms{Michael R.} \snm{Kosorok}\ead[label=e3]{kosorok@unc.edu}}

\affiliation[*]{The University of North Carolina at Chapel Hill}
\affiliation[**]{University of Haifa}

\address{Sayan Dasgupta \& Michael R. Kosorok \\Department of Biostatistics
\\The University of North Carolina at Chapel Hill\\Chapel Hill, NC 27599, U.S.A.\\ \printead{e1}\\\printead{e3}}

\address{Yair Goldberg\\Department of Statistics
\\The University of Haifa\\ Mount Carmel, Haifa 31905, Israel\\ \printead{e2}}


\runauthor{Dasgupta, Goldberg and Kosorok}
\end{aug}

\begin{abstract}
We develop an approach for feature elimination in statistical learning with kernel machines, based on recursive elimination of features. We present theoretical properties of this method and show that it is uniformly consistent in finding the correct feature space under certain generalized assumptions. We present four case studies to show that the assumptions are met in most practical situations and present simulation results to demonstrate performance of the proposed approach.
\end{abstract}


\end{frontmatter}

\section{Introduction}\label{sec:intro}
\defcitealias{SVR}{SC08}
With recent advancement in data collection and storage, we have large amounts of information at our disposal, especially with respect to the number of explanatory variables or `features'. When these features contain redundant or noisy information, estimating the functional connection between the response and these features can become quite challenging, and that often hampers the quality of learning. One way to overcome this is by finding a smaller set of features or explanatory variables that can perform the learning task sufficiently well. 

In this paper, we discuss feature elimination in statistical learning with kernel machines. Kernel machines (KM) are a class of learning methods for pattern analysis and regression, under transformations of the input feature space, of which the linear support vector machine (SVM) is the simplest case. In general, the term `kernel machine' is reserved for the more general version of the SVM problem with non-linear transformation of the feature space. The popularity of these algorithms is motivated by the fact that these are easy-to-compute techniques that enable estimation under weak or no assumptions on the distribution \citep[see][]{SVR}. Kernel machines, which we review in Section~\ref{sec:summary}, are a collection of optimization algorithms that attempt to minimize a regularized version of the empirical risk over some reproducing kernel Hilbert space (RKHS) of functions (referred to as $H$) for a given loss function $L$. The standard KM decision function typically utilizes all the input features. However, the prediction quality of these methods often suffers under high noise-to-signal ratio, even if the input space is only moderately high dimensional. In Section \ref{simu_lasso}, we present an example (see Table \ref{tab:class}) for a non-linear classification with only ten features, of which only two are relevant. We see that applying a meaningful feature selection method there can cut classification error from $18\%$ to about $6\%$. It is thus a very important task to be able to select the correct feasible set of input features on which the learning can be applied. 

Multiple methods have been proposed for the linear version of the problem, that is, when the assumed functional form of the decision rule is linear. For example, many \emph{embedded methods}\footnote{methods that construct the learning algorithm in a way to include feature elimination as an in-built phenomenon.} with different modifications  have been proposed, like redefining the SVM training to include sparsity in \citep{Weston2003}, using the $l_1$ penalty as in \citet{Bradley1998,Zhu2003}, the SCAD penalty in \citep{Zhang2006scad}, the $l_{q}$ penalty \citep{Liu2007lq} or the elastic net \citep{Wang2006}. Although these methods have strong theoretical guarantees, they are relevant only in linear SVM, and become ineffectual in the framework of RKHSs with non-linear kernels (like the Gaussian RBF kernel).  The widely applicable linear SVM, the most popular and well-known of the general class of KM problems, has been the focus of most of the prevalent feature selection techniques. Non-linear versions of the algorithm (the kernel machine setup) have however become increasingly important these days, and many statistical learning problems explicitly depend on functional relationships that are strictly non-linear in nature - for example, in protein classification \citep[see][]{Leslie2004}, and in image classification \citep[see][]{Chapelle1999}, etc. Hence feature selection for kernel machines is the key focus for us in this paper.

A few techniques do exist that can be effectively catered to the kernel machines framework or the non-linear SVMs. For example, \citet{Guyon2002} developed a \emph{wrapper}\footnote{methods that use the learning method itself to score feature subsets.}-based backward elimination procedure by recursively computing the learning function, known widely as recursive feature elimination (RFE). Although RFE was developed as an off-the-shelf technique for linear SVMs, the authors included an analogous formulation for the non-linear transformed space as well. The RFE algorithm performs a recursive ranking of a given set of features. At each recursive step of the algorithm, it calculates the change in the RKHS norm of the estimated function after deletion of each of the features remaining in the model, and removes the one with the lowest change in such norm, thus performing an implicit ranking of features. 
A number of approaches have been developed inspired by RFE \citep[see][]{Rakotomamonjy2003,Tang2007,Mundra2010}. The idea of feature selection in a backward recursive method that is essentially the founding principle behind RFE, has been studied extensively in the Bioinformatics and Computer Science literature \citep[see for example][]{Zhang2006recursive, Aksu2010, Aksu2012}. RFE has been used for feature selection in many recent applications \citep[see for example][]{Hu2010,Hidalgo2013, Louw2006}. Other selection methods also exist as can be seen in \citet{Weston2001,Chapelle2002}. Recently, a new multistage \emph{embedded} optimization method has been proposed \citep[see][]{Allen2013}. However, the key drawback of these methods is that their theoretical properties have never been studied rigorously.

A key reason behind this lack of theory is the absence of a well-established framework for building, justifying, and collating the theoretical foundation of such a feature elimination method. This paper aims at building such a framework and modifying RFE to create a recursive technique that can be validated as a theoretically sound procedure for feature elimination in kernel machines. Our main contributions include: \\
(1) We \textbf{develop a theoretical framework} that can validate feature elimination in KMs. For example, the KM algorithm attempts to minimize the empirical regularized risk within an RKHS defined on the initial input space $\mathcal{X}$. Hence, one important task is to redefine $H(\cdot)$ on any lower dimensional domain so that it retains its RKHS properties. The basis for the theory on RFE depends heavily on correctly specifying these pseudo-subspaces. \\
(2) We \textbf{modify the criterion for deletion and ranking of features} from \citeauthor{Guyon2002}'s RFE. The ranking of the features here is based on the lowest difference observed in the regularized empirical risk after removing each feature from the existing model. This is done to enable theoretical consistency.\\
(3) We establish \textbf{asymptotic consistency of the modified RFE algorithm} in finding the `correct' feature space. We give necessary conditions for achieving consistency, and show using counterexamples that these conditions are indeed necessary. We believe these are some of the first theoretical results on feature selection in kernel machines.\\
(4) We discuss the applicability of our methods in \textbf{four different learning problems, including protein and image classification}. \\
(5) We extend the scope of our algorithm for situations when the \textbf{dimensionality of the input space is allowed to grow with the sample size}, and establish a range of rates of such growth that can be allowed to still guarantee consistency.

The paper is organized as follows: In Section \ref{sec:summary}, we present a short summary of the problem, the proposed feature elimination algorithm for kernel machines, and the main theoretical result of the article. In Section \ref{sec:valid}, we discuss consistency of the algorithm in the simpler yet practical settings of nested or dense models. Four case studies are discussed in depth in Section \ref{sec:cs}. In Section \ref{sec:disc_assump}, we relax earlier assumptions to allow us to establish consistency in more complex functional spaces. In Section \ref{sec:theory}, we prove our main results under the most general setting. In Section \ref{sec:large_p} we discuss the scope of the algorithm when the size of the covariate space $d$ grows with sample size $n$. In Section \ref{sec:simu}, we provide simulation results to demonstrate how risk-RFE works and how it can be used in intelligent selection of features, and compare it with different penalized methods for feature selection. A~discussion is provided in Section \ref{sec:dis}, and additional materials along with the detailed proofs are given in the Appendix. Some additional resources and the necessary software codes are given in the \ref{sec:suppA}.

\section{The Risk-Recursive Feature Elimination Algorithm (risk-RFE)}\label{sec:summary} In this section, we summarize the main findings of the paper. We first describe the relevant problem briefly, along with its mathematical formulation, and then follow it up by our proposed risk-RFE algorithm and the main consistency result for it.

\subsection{The problem description}\label{sec:prob} Kernel machines are a set of supervised learning methods that has become a very useful tool in statistical learning problems in both classification and regression, especially when transformation of the input space becomes necessary to derive an optimal rule. Given training data  $D=\{(X_{1},Y_{1}),\dots,(X_{n},Y_{n})\} \in (\mathcal{X} \times \mathcal{Y})^{n}$, the goal is to estimate a rule that can be used to predict $Y_i$ for a given input feature vector $X_i$. In kernel machines, this is done by minimizing a regularized version of the empirical risk (for a given loss function $L$) of functional rules obtained from special functional spaces, called the reproducing kernel Hilbert space (RKHS). An RKHS $H$ is typically represented by a bilinear function $k(\cdot,\cdot)$, and for a given transformation $\phi$ of the feature space, the appropriate RKHS $H$ is the one with kernel $k$ satisfying $k(x_1,x_2)=\left\langle \phi(x_1),\phi(x_2) \right\rangle_{H}$. This property, called the kernel trick, circumvents the need for explicit knowledge of the feature map $\phi$, and the non-linear decision function can be represented in terms of the function $k(\cdot,\cdot)$.

The linear SVM for binary classification with the hinge loss $L_{HL}(X,Y,f(X))=\max(1-Yf(X),0)$ is the most popular version of the kernel machine algorithm, but for the untransformed feature space. It has been extensively studied, under various variable selection methods (especially the $L_p$ penalized embedded forms of the algorithm as they are easily interpretable in the linear case). However, feature selection in general kernel machines is still a relatively new area of research, and our goal here is to lay the foundation of a feature selection method that is technically able to consistently estimate the correct feature space in these scenarios.

\subsection{Mathematical formulation} The notations and the type of oracle bounds used in this paper will closely follow \citet{SVR} \citepalias[hereafter abbreviated][]{SVR}. Consider the measurable space $(\mathcal{X},\mathcal{A},P_{\mathcal{X}})$ such that $\mathcal{X} \subseteq B \subset \mathbb{R}^d$ is a valid metric space, where $B$ is an open Euclidean ball centered at $0$. Let $\mathcal{Y}$ be a closed subset of $\mathbb{R}$ and $P$ be a measure on $\mathcal{X}\times\mathcal{Y}$, such that $P_{\mathcal{X}}$ is a restriction of $P$ on $\mathcal{X}$. We start by defining the kernel machine algorithm in its most general forms.

\textsc{Kernel machine (KM) :} Let $L: \mathcal{X}\times\mathcal{Y}\times\mathbb{R}$ $\mapsto$ $[0,\infty]$ be a convex, locally Lipschitz continuous and measurable loss function and $H$ be a separable RKHS of a measurable kernel $k$ on $\mathcal{X}$, and fix a $\lambda>0$. The \emph{general KM solution} is the function $f_{P,\lambda,H} \in H$ that satisfies 
\begin{align*}
f_{P,\lambda,H}= \underset{f \in H}{\arg\min}\lambda\left\|f\right\|^{2}_{H}+\mathcal{R}_{L,P}(f).
\end{align*}
For the observed data $D$, the \emph{empirical KM decision function} is then given as 
\begin{align}\label{eq:svm}	
f_{D,\lambda,H}=\underset{f \in H}{\arg\min}\lambda\left\|f\right\|^{2}_{H}+\mathcal{R}_{L,D}(f).
\end{align}

\begin{rem}
The loss function $L$ is convex if $L(x,y,\cdot)$ is convex for every $x\in\mathcal{X}$ and $y\in\mathcal{Y}$. It is also locally Lipschitz continuous if for every $a>0$, $\sup_{x\in\mathcal{X},{y\in\mathcal{Y}}}\left|L(x,y,s)-L(x,y,\acute{s})\right| < c_{L}(a)\left|s-\acute{s}\right|,\: s,\acute{s}\in [-a,a]$ for a given local constant $c_{L}(\cdot)$. Note that the results developed here are equally valid for regression under certain general assumptions on $\mathcal{Y}$. 
\end{rem}
\begin{rem}
The L-risk of the measurable function $f$ is given as $\mathcal{R}_{L,P}(f)=E_{P}[L(X,Y,f(X)]$. The Bayes risk $\mathcal{R}^{\ast}_{L,P}$ is defined as $\inf_{f}{\mathcal{R}_{L,P}(f)}$, where the infimum is taken over the set of all measurable functions,
	$\mathcal{L}_{0}(\mathcal{X})=\{f:\mathcal{X}\mapsto\mathbb{R},\:\: f \text{ is measurable}\}$.
A function $f^{\ast}_{P}$ that achieves this infimum is called a Bayes decision function. Now let $f_{P,\mathcal{F}} = \arg\min_{f \in \mathcal{F}} E_{P}[L(X,Y,f(X)] = \arg\min_{f \in \mathcal{F}} \mathcal{R}_{L,P}(f)$ be the minimizer of infinite-sample risk within $\mathcal{F}$. We denote this minimal risk as $\mathcal{R}^{\ast}_{L,P,\mathcal{F}} = \mathcal{R}_{L,P}(f_{P,\mathcal{F}})$.
\end{rem}

\begin{rem}
The kernel $k$ of the RKHS H is unique, a real-valued symmetric function $k:\mathcal{X}\times\mathcal{X} \mapsto \mathbb{R}$. The kernel $k$ has the reproducing property that $f(x)=\left\langle f, k(\cdot,x)\right\rangle_H$ for all $f \in H$, and all $x \in \mathcal{X}$, where $\left\langle \cdot,\cdot\right\rangle_H$ is the inner product induced by $H$. Moreover, we also have $k(\cdot,x) \in H$, for all $x \in \mathcal{X}$. 
\end{rem}

%

\subsection{The feature elimination algorithm}\label{sec:feat_algo} Limitations of \citeauthor{Guyon2002}'s RFE as a margin-maximizing feature elimination were studied explicitly in \citet{Aksu2010}. 
Hence as opposed to \citeauthor{Guyon2002}, who used the Hilbert space norm $\lambda\|f\|^{2}_{H}$ to eliminate features recursively, we use the entire objective function (the regularized empirical risk) for deletion.
For a given loss function $L$ and probability measure $Q$, define the regularized empirical risk as \[\mathcal{R}^{\text{reg},\lambda}_{L,Q,\mathcal{F}}\left(f\right)= \lambda\left\|f\right\|^{2}_{\mathcal{F}} + \mathcal{R}_{L,Q}\left(f\right).\] Also define the restricted space $\mathcal{F}^{J}$ as follows:
\begin{defn}\label{defn:fj}
Let $J$ be a set of indices $J \subseteq \{1,2,..,d\}$. Then for a given functional space $\mathcal{F}$, define
$\displaystyle\mathcal{F}^{J} = \{g:g=f\circ\pi^{J^{c}} \quad , \forall f \in \mathcal{F}\}$, where $\pi^{J^{c}}$ is the projection map that takes element $x\in \mathbb{R}^d$ and maps it to $x^{J}\in\mathbb{R}^{d}$, by substituting elements in $x$ indexed in the set $J$, by zero.
\end{defn}

\begin{rem}
Note that we can subsequently define the space $\mathcal{X}^{J}=\{\pi^{J^{c}}(x)\::\:x \in\mathcal{X}\}$. Thus the above formulation allows us to create lower dimensional versions of a given functional space $\mathcal{F}$.
\end{rem}
We are now ready for our feature selection method. The risk-RFE algorithm, defined for the parameters $\{\lambda_n, \delta_n\}$ is given as:
\begin{algo}[risk-RFE]\label{algo:svm}
Start off with $J \equiv [\cdot]$ empty and let $Z \equiv [1,2,...,d]$.\\

STEP 1: In the $k^{th}$ iteration, choose feature $i_{k}\in Z \setminus J$ which minimize
\begin{align}\label{eq:opt_step}
 \mathcal{R}^{\text{reg},\lambda_n}_{L,D,H^{J \cup \{i\}}}\left(f_{D,\lambda_n,H^{J \cup \{i\}}}\right) -  \mathcal{R}^{\text{reg},\lambda_n}_{L,D,H^{J}}\left(f_{D,\lambda_n,H^{J}}\right),
\end{align}
STEP 2: Update $J = J \cup \{i_{k}\}$. Go to STEP 1.\\

Continue this until the difference $\displaystyle\min_{i \in Z \setminus J} \mathcal{R}^{\text{reg},\lambda_n}_{L,D,H^{J \cup \{i\}}}\left(f_{D,\lambda_n,H^{J \cup \{i\}}}\right) -  \mathcal{R}^{\text{reg},\lambda_n}_{L,D,H^{J}}\left(f_{D,\lambda_n,H^{J}}\right)$ becomes larger than a pre-determined quantity $\delta_{n}$, and output $J$ as the set of indices for the features to be removed from the model.
\end{algo}
The optimal choice of the parameters $\lambda_n$ and $\delta_n$ are discussed in Section \ref{sec:consis}.

\subsection{Number of features removed in each iteration of RFE} In Algorithm \ref{algo:svm} above, we have discussed removing only one feature at each iteration. Similarly, one can also consider removing multiple features (say $k$) in a single iteration. In that case those $k$ indices are removed that produce the $k$ smallest values of the objective function \eqref{eq:opt_step} (given above) in that iteration. This number can also be defined adaptively, such that different numbers of features are removed in different iterations of the algorithm. For simplicity, we have set it to $1$ for our theory, but in numerical simulations we have often defined it adaptively to speed up computations.


\subsection{Heuristics on why RFE is consistent}\label{sec:featel_svm} 
Consider risk minimization in general, and let our goal be to find a solution $f\in\mathcal{F}$ that minimizes a given empirical criterion (like `regularized empirical' risk in kernel machines). If it so happens that the minimizer of  the infinite-sample risk resides in a space spanned by a lower dimensional subspace of $\mathcal{X}$ (say $\mathcal{X}^{\ast}$), then it may actually suffice to find this empirical minimizer over the restriction of $\mathcal{F}$ on $\mathcal{X}^{\ast}$. To avoid overfitting, this indeed becomes necessary. The need for defining the lower dimensional adaptations of a given functional class in Definition \ref{defn:fj} in section \ref{sec:feat_algo} arises from this observation itself. Hence the motivation for our algorithm stems from the following heuristic: if any feature is superfluous for the given problem, then given all other features in the model, its contribution to the functional relationship between the output variable and the feature space should only be due to random fluctuations and therefore small. Hence the incremental risk associated with a solution in the subspace defined by ignoring this surplus feature, when compared to the solution in the original feature space, should be minimal. 

\subsection{Consistency results for RFE}\label{sec:consis} The main result of this paper states that algorithm \ref{algo:svm} presented in section \ref{sec:feat_algo} is consistent in finding the correct feature space in spaces admitting the nestedness or denseness property. We thus assume Condition \ref{cond:1} below (see Section \ref{sec:valid} for a detailed discussion of Condition \ref{cond:1} and definitions of nestedness and denseness). We further assume $d$, the dimension of $\mathcal{X}$, and $d_0$, the number of relevant features, to be fixed constants.  Also let $J_{\ast}$ be the maximal set satisfying the condition $\mathcal{R}_{L,P,\mathcal{F}}^{\ast}=\mathcal{R}_{L,P,\mathcal{F}^{J_{\ast}}}^{\ast}$.

 
\begin{cond}\label{cond:1}\ \\
The functional space admits the nestedness or denseness property.
\end{cond}
\begin{thm}\label{thm:main2} 
Let $L$ be a convex locally Lipschitz continuous loss function satisfying $L(x,y,0) \leq 1$ for all $(x,y) \in \mathcal{X}\times\mathcal{Y}$. Let $H$ be the separable RKHS of a measurable kernel $k$ on $\mathcal{X}$ with $\|k\|_{\infty}\leq 1$. Assume, for fixed $n\geq1$, there exist constants $a\geq1$ and $p\in (0,1)$ such that the entropy condition \eqref{cond:ent} given below holds. Let $\{\lambda_{n}\}\in [0,1]$ be such that $\lambda_{n} \rightarrow 0$ and $\displaystyle\lim_{n\rightarrow\infty}\lambda_{n}n=\infty$. Also assume that there exists $c>0$ and $\beta \in (0,1]$ such that $A_{2}^{J_{\ast}}(\lambda)\leq c\lambda^{\beta}$ for all $\lambda \geq 0$. 

Then for $\delta_n=\epsilon_0-O(n^{-\frac{\beta}{2\beta+1}})$, the following statements hold:
\begin{enumerate}
	\item The Recursive Feature Elimination Algorithm for kernel machines, defined for $\{\delta_{n},\lambda_{n}\}$ given above, will find the correct lower dimensional subspace of the input space ($\mathcal{X}^{J_{\ast}}$) with probability tending to $1$.\label{itm:part1_svr}
	\item The function chosen by the algorithm achieves the minimal risk within the original RKHS $H$ asymptotically.\label{itm:part2_svr}
\end{enumerate}
\end{thm}
We need to define the entropy condition on the complexity of the RKHS $H$. Before that, note that for a given metric space $(T,d)$ and for any integer $n\geq 1$, the $n-$th entropy number of $(T,d)$ is defined as
\begin{align*}	
	e_{n}(T,d):=\inf\left\{\epsilon>0\::\:\exists s_{1},\dots,s_{2^{n-1}}\in T\:\text{such that}\:T\subset \displaystyle\bigcup_{i=1}^{2^{n-1}}B_{d}(s_{i},\epsilon)\right\},
\end{align*}
where $B_d(s,\epsilon)$ is the ball of radius $\epsilon$ centered at $s$, with respect to the metric $d$. If $S:E\mapsto F$ is a bounded linear operator between normed spaces $E$ and $F$, we write $e_{n}(S)=e_{n}(SB_{E},\|\cdot\|_{F})$, where $B_E$ is the unit ball in $E$.

Then the \emph{entropy condition} \eqref{cond:ent} is given as
\begin{align}\label{cond:ent}
\mathbb{E}_{D_{\mathcal{X}}\sim P^{n}_{\mathcal{X}}}e_{i}\left(id:H\mapsto L_{\infty}(D_{\mathcal{X}})\right) \leq ai^{-\frac{1}{2p}},\: i \geq 1,
\end{align}
where $\mathbb{E}_{D_{\mathcal{X}}\sim P_{\mathcal{X}}^{n}}$ is defined as the expectation with respect to the product measure  $P_{\mathcal{X}}^{n}$ for data $D_{\mathcal{X}}\equiv \{\mathcal{X}_1,\dots,\mathcal{X}_n\}$ being i.i.d. copies of $\mathcal{X}\sim P_{\mathcal{X}}$, and $a$ and $p$ are fixed constants.
\begin{rem}
The proof of this theorem in a more general setting is given in Section \ref{sec:main2_proof}.
\end{rem}

\begin{rem}
Conditions $L(x,y,0) \leq 1$, and $\|k\|_{\infty}\leq 1$ for the kernel `$k$' in Theorem \ref{thm:main2} are assumed for simplicity and equivalent conditions such as $L(x,y,0) \leq M$ and $\|k\|_{\infty}\leq k_{\text{sup}}$ for constants $M,\:k_{\text{sup}}>1$ will suffice to guarantee the desired result.
\end{rem}
\begin{rem}
Note that the approximation error $A^{H}_2(\lambda)$ or the bias is defined as the difference between the regularized risk of $f_{P,\lambda,H}$ and the minimum risk achieved within $H$, and is given by $A_{2}^{H}(\lambda)=\mathcal{R}^{\text{reg},\lambda}_{L,P,H}\left(f_{P,\lambda,H}\right) - \mathcal{R}^{\ast}_{L,P, H}$. Consistency follows even in absence of the exponential bound on the approximation error. However allowing such a bound allows us to derive explicit rates for our algorithm. Also note that we actually need  $A_{2}^{J}(\lambda)\leq c\lambda^{\beta}$ for any $J\subseteq J_{\ast}$ in general, but when nestedness or denseness holds, it actually suffices to have the condition hold for $J=J_{\ast}$. 
\end{rem}
The proof is postponed to Section \ref{sec:theory}.

\section{RFE in nested or dense models}\label{sec:valid} Here we discuss the scope of our feature elimination algorithm when the functional space admits certain nice properties like nestedness or denseness. We begin by defining these properties and presenting important examples of such spaces. We then discuss our inherent assumption of `existence of a null model' under these setups, and show how this inherently translates into the idea of feature elimination through risk-RFE.  

\subsection{Nested spaces in risk minimization}\label{sec:nest_space} In risk minimization, we say $\mathcal{F}$ admits the nested property, if for a pair of index sets $J_{1}, J_{2} \in \{1,2,\dots,d\}$ with $J_{1}\subseteq J_{2}$, we have $\mathcal{F}^{J_{2}}\subseteq\mathcal{F}^{J_{1}}$. This translates to admitting nested inequalities of the form $\mathcal{R}_{L,P,\mathcal{F}^{J_{1}}}^{\ast}\leq\mathcal{R}_{L,P,\mathcal{F}^{J_{2}}}^{\ast}$. One simple example is the linear space,
$\mathcal{F}=\big\{f(x_{1},\dots,x_{d})=\sum_{i}a_{i}x_{i}\::\:|a_{i}|\leq M,\: M<\infty \big\}$. 

In general, RKHSs need not be nested within each other. 
The question is when can naturally occurring RKHSs be nested? We will see below that dot-product kernels actually do have this property. 

\begin{lem}\label{lem:dot_nested}
Dot product kernels produce nested RKHSs.
\end{lem}

The proof is given in the \ref{sec:suppA}. Dot product kernels (e.g., linear kernels) appear quite regularly in the formulation of KM problems. Other kernels might display the nested property. We will discuss the usefulness of this in section~\ref{sec:nullmodel}.

\subsection{Dense spaces in risk minimization}\label{sec:dense_space} We say that $\mathcal{F}$ admits the denseness property, if it is dense in a functional class that admits the nested property (for example, the space of bounded measurable functions $\mathcal{L}_{\infty}(\mathcal{X})$ or the space of continuous and bounded functions $\mathcal{C}(\mathcal{X})$). Note that all universal kernels produce RKHSs that are dense in $\mathcal{C}(\mathcal{X})$, and also in $\mathcal{L}_{\infty}(\mathcal{X})$ if the loss function is convex and locally Lipschitz continuous. All non-trivial radial kernels (e.g., Gaussian RBF kernel) share this property as well \citep[see][]{Micchelli06}, and hence this is quite a typical framework for KM problems. 
%

\subsection{Existence of a null model}\label{sec:nullmodel} Existence of a null model essentially means that there exists an index set $J_{\ast}$ such that,
\begin{align}\label{eqn:assump_nested}
\mathcal{R}_{L,P,\mathcal{F}}^{\ast}=\mathcal{R}_{L,P,\mathcal{F}^{J_{\ast}}}^{\ast}.
\end{align}

\begin{rem}
Note that the above does not claim the uniqueness of $J_{\ast}$. Rather we say that for any set of covariates with the above property \eqref{eqn:assump_nested}, there always exists a maximal set in terms of it. Also note that $J_{\ast}$ can be empty. Here empty means that $f_{\ast}$ does not reside in any proper restricted subspace of $\mathcal{F}$ and thus we would need the entire space $\mathcal{F}$ for the optimization. 
\end{rem}
In spaces which admit the nestedness property, existence of a null model is equivalent to saying that there exists a minimizer of infinite-sample risk in $\mathcal{F}$, which also lives in $\mathcal{F}^{J_{\ast}}$. This then trivially implies $\mathcal{R}_{L,P,\mathcal{F}^{J}}^{\ast}=\mathcal{R}_{L,P,\mathcal{F}^{J_{\ast}}}^{\ast}$ whenever $J\subseteq J_{\ast}$. If $\mathcal{F}$ is now dense in a functional space $\mathcal{G}$ admitting the nested property (for eg. $ \mathcal{L}_{\infty}(\mathcal{X})$), then by Lemma \ref{lem:fandfj} of \ref{sec:suppA}, $\mathcal{F}^J$ is dense in $\mathcal{G}^J$ for any $J\in\{1,2,\dots,d\}$. Hence `denseness' does not necessarily imply `nestedness', but we do have the `almost nested' property in the sense that any function $g \in \mathcal{F}^{J_{2}}$ can be well approximated by a sequence of functions $\{f_n\} \in \mathcal{F}^{J_{1}}$ for $J_{1}\subseteq J_{2}$. This actually implies \eqref{eqn:assump_nested} (given above) for any $J\subseteq J_{\ast}$. Then in the finite $d$ setting, there exists an $\epsilon_{0}>0$, such that $\mathcal{R}^{\ast}_{L,P,\mathcal{F}^{J_{\circ}}}\geq \mathcal{R}^{\ast}_{L,P,\mathcal{F}^{J_{\ast}}}+\epsilon_{0}$ holds whenever $J_{\circ} \nsubseteq J_{\ast}$. 
\begin{rem}
The `nested structure' is essentially different from the nested model setup in \citet{Tsybakov2004}. \citet{Tsybakov2004} started with a pre-decided nested sequence of classifier sets (or models) and obtained a solution from each of these classifier sets. But here we have a graph of nested models that can include many subtrees in the sense that in every intermediate step, we are presented with multiple models within the parent model. We select the best classifier from each of these models and opt for the one among them obtaining the best performance.
\end{rem}

\section{Case studies}\label{sec:cs} In this section we show the validity of our results in many practical cases of risk minimization by discussing the results in some known settings.

\subsection{Case Study 1: Linear Regression as a SVM problem with zero regularization}\label{sec:cs1} 
In a linear regression model, the functional relationship is expressed as $y=\left\langle \alpha, x\right\rangle + b_{0}$, where $\left\langle\alpha, x\right\rangle$ denotes the Euclidean inner product of vectors $\alpha$ and $x$, and $b_{0}$ is the bias. The prediction quality of this model can be measured by the squared-error loss function $L_{LS}$ given as $L_{LS}(x,y,f(x))=(f(x)-y)^{2}$. Our goal is to find linear weights $\widehat{\alpha}$ and $\widehat{b}_{0}$ for the observed data $D$ that minimize the empirical risk. We assume that the input space $\mathcal{X}\subseteq B\subset \mathbb{R}^{d}$. We further assume that $\mathcal{Y}\subset\mathbb{R}$ is a closed set. The functional space $\mathcal{F}_{\text{lin}}$ is given by
$\mathcal{F}_{\text{lin}}=\{f_{\alpha,b_{0}}:f_{\alpha,b_{0}}(x)=\left\langle \alpha, x\right\rangle + b_{0},\:(\alpha,b_{0}) \in\mathbb{R}^{d+1},\:\|(\alpha,b_{0})\|_{\infty}\leq M,\:\text{\emph{for some}}\: M<\infty\}$.
We can now observe that the regularity conditions
required for consistency of the recursive algorithm hold for this problem. The least squares loss function $L_{LS}$ is convex and is locally Lipschitz continuous when $\mathcal{Y}$ is compact.
The linear space $\mathcal{F}_{\text{lin}}$ is also an RKHS with Euclidean inner product as its kernel function, and it can be shown that it satisfies all the necessary regularity conditions.  
We can assume an exponential bound on the average entropy number, and in fact much stronger bounds can be obtained for the $\epsilon$-entropies of the linear functional class \citep[see for example][]{Zhang2002,Williamson2000}.
%
The risk-RFE procedure presented in this paper translates in the linear regression case to a non-parametric backward selection method based on the value of the `average sum of squares of error' or $R^{2}$. Note that under restrictive distributional assumptions on the output vector $\mathcal{Y}$ in a non-parametric setup, the idea of using penalized versions of $\log R^{2}$ like AIC, AICc or BIC are well accepted methodologies for model selection. 
%

It must be noted that in linear regression risk-RFE can only be used when $d\leq n$ as identifiability becomes an issue otherwise. \citet{Fan2007} discussed the setup of $d>n$ in detail, and we discuss this premise briefly in section \ref{sec:dis}, and omit any further mention here.
 

\subsection{Case Study 2: Support vector machines with a Gaussian RBF kernel}\label{sec:cs2} Here we provide a brief review of the application of risk-RFE in the classic SVM premise for classification using a Gaussian RBF kernel. Assume that $\mathcal{Y}=\{1,-1\}$. We want to find a function $f:\mathcal{X}\mapsto \{1,-1\}$ such that for almost every $x\in\mathcal{X}$, $P(f(x)=Y\big|\mathcal{X}=x) \geq 1/2$.
In this case, the desired function is the Bayes decision function $f^{\ast}_{L,P}$ with respect to the loss function $L_{BC}(x,y,f(x)) = 1\{y\cdot \text{sign}(f(x))\neq 1\}$. In practice, since $L_{BC}$ is non convex, it is usually replaced by the hinge loss function $L_{HL}(x,y,f(x)) = \max\{0, 1-yf(x)\}$. For SVMs with a Gaussian RBF kernel, we minimize the regularized empirical criterion $\lambda\|f\|^{2}+\frac{1}{n}\sum_{i=1}^{n}\max\{0, 1-y_{i}f(x_{i})\}$ for the observed sample $D=\{(x_{1},y_{1}),\dots,(x_{n},y_{n})\}$ within the RKHS $H_{\gamma}(\mathcal{X})$ with the kernel $k_{\gamma}$ defined as $k_{\gamma}(x,y)=e^{-\gamma^{2}\|x-y\|^{2}_{2}}$, where $\sigma:=1/\gamma$ is called the width of the kernel $k_{\gamma}$. 

\begin{lem}\label{lem:rbf_delta}
For classification using support vector machines with a {G}aussian RBF kernel, the RFE defined for $\delta=\epsilon_0-O(n^{-\frac{\beta}{2\beta+1}})$ where $\beta=\frac{\alpha}{\alpha+1}$, where $0<\alpha<\infty$ is the geometric noise exponent of $P$ on $\mathcal{X}\times \{-1,1\}$.
\footnote{For a discussion on the geometric noise exponent we refer our readers to \citet{Steinwart07}.}
\end{lem}

Lemma \ref{lem:rbf_delta} gives us a precise characterization of $\delta_n$ in this setting in terms of the geometric noise exponent of $P$ on $\mathcal{X} \times \{1,-1\}$. The proof is given in the Appendix .

\subsection{Case Study 3: Protein classification with Mismatch String kernels}\label{sec:cs3} A very fundamental problem in computational biology is the classification of proteins into functional and structural classes based on homology of protein sequence data. A new class of kernels called the mismatch string kernels, are increasingly being used with kernel machines	 in a discriminative approach to the protein classification problem. These kernels measure sequence similarity based on shared occurrences of $k$ length subsequences, counted with up to $m$ mismatches. This is again a typical classification problem, where $\mathcal{Y}=\{1,-1\}$ and the hinge loss function $L_{HL}(x,y,f(x)) = \max\{0, 1-yf(x)\}$ is again used as the surrogate loss. The $(k,m)$ mismatch kernel \citep[see][for details]{Leslie2004} is based on a feature map from the space of all finite sequences from an alphabet $\mathcal{A}$ (with $\mathcal{C}(A)=l$) to ${\mathbb{Z}_{\geq 0}}^{l^k}$, where $l^k$ denotes the dimensions spanned by the set of $k$-length subsequences (`$k$-mers') from $\mathcal{A}$. For a fixed $k$-mer $\alpha=a_{1}a_{2} \dots a_{k}$, with each $a_{i}$ a character in $\mathcal{A}$, the $(k,m)$-neighborhood generated by $\alpha$ is the set of all $k$-length sequences $\beta$ from $\mathcal{A}$ that differ from $\alpha$ by at most $m$ mismatches. We call this set $N_{(k,m)}(\alpha)$. 

The feature map $\Phi_{(k,m)}$ for a $k$-mer $\alpha$ is defined as $\Phi_{(k,m)}(\alpha)=(\phi_{\beta}(\alpha))_{\beta \in \mathcal{A}^{k}}$, where $\phi_{\beta}(\cdot)$ is a indicator function such that, $\phi_{\beta}(\alpha)=1$ if $\beta \in N_{(k,m)}(\alpha)$, and $0$ otherwise. Then for a sequence $x$ of any length, the feature map $\Phi_{k,m}$ is defined as follows:
\begin{align*}
\Phi_{(k,m)}(x)=\displaystyle\sum_{k-\text{mers }\alpha\text{ in }x} \Phi_{(k,m)}(\alpha),
\end{align*}
that is, we extend the feature map additively by summing the feature vectors for all the $k$-mers in $x$. The $(k,m)$-mismatch kernel $K_{(k,m)}(x,y)$ is then the Euclidean inner product in the space of feature vectors:
\begin{align*}
K_{(k,m)}(x,y)=\left\langle \Phi_{(k,m)}(x), \Phi_{(k,m)}(y) \right\rangle.
\end{align*}
For $m=0$, we retrieve the $k$-spectral kernel. The kernel can be further normalized as
\begin{align*}
K_{(k,m)}^{\text{norm}}(x,y)=\frac{K_{(k,m)}(x,y)}{\sqrt{K_{(k,m)}(x,x)}\sqrt{K_{(k,m)}(x,y)}}.
\end{align*}
Feature selection in the context of protein classification is conducted on the $k$-mers obtained from a protein sequence instead of the original one \citep[see][]{Leslie2004,Iqbal2014}. It can be observed that the RKHS $H$ produced by the string kernel is finite dimensional and hence satisfies the regularity conditions on the RKHS trivially, and hence, the coordinates of the transformed space (the $k$-mers) can be used directly for feature selection. The problem reduces down to feature selection in linear SVMs (produced by the Euclidean inner product), and the applicability of recursive feature selection becomes clear in the context of the discussions we had in case studies \ref{sec:cs1} and \ref{sec:cs2}.   

\subsection{Case Study 4: Image classification with $\chi^2$ kernel}\label{sec:cs4} Indexing or retrieving images is one of the main challenges in pattern recognition problems. Using color histograms as an image representation technique is useful because of the reasonable performance that can be obtained in spite of their extreme simplicity \citep[see][]{Swain1991}. Image classification using their histogram representation has become a popular option in many such settings, and the kernel machine approach is considered a good classification technique here
 \citep[see][]{Chapelle1999}. 

Selecting the kernel is important in these problems, and generalized RBF kernels of the form $K^{d-\text{RBF}}_{\rho}(x,y)=e^{-\rho d(x,y)}$ have been proven to be useful for classification in this context. In case of images as input, the histograms produced generate discrete densities and suitable comparison functions like the $\chi^2$ function are preferred over the $L_{2}$ norm that generates the Gaussian RBF kernel, and has been used extensively for histogram comparisons \citep[][]{Schiele1996}.
The $\chi^2$ distance is given as $d_{\chi^2}(x,y)=\sum_{i}\frac{(x_i-y_i)^2}{x_i+y_i}$, and hence the $\chi^2$ kernel has the form,
\begin{align*}
K^{\chi^2-\text{RBF}}_{\rho}(x,y)=e^{-\rho \sum_{i}\frac{(x_i-y_i)^2}{x_i+y_i}}.
\end{align*}
If the regularity conditions given before Theorem \ref{thm:main2} in section \ref{sec:consis} are satisfied in this setup, that will establish consistency for risk-RFE here. Note that we have already established the conditions for hinge loss $L_{HL}$ in the case study \ref{sec:cs2}. Kernel $K^{\chi^2-\text{RBF}}_{\rho}$ is continuous, and the input space is separable, hence separability of $H^{\chi^2-\text{RBF}}_{\rho}$ follows from Lemma $4.33$ of \citetalias{SVR}. It also follows that $\|K^{\chi^2-\text{RBF}}_{\rho}\|_{\infty}\leq 1$. Since the input space $\mathcal{X}$ can be included in a Euclidean ball and the kernel $K^{\chi^2-\text{RBF}}_{\rho}$ is infinitely many times differentiable, Theorem $6.26$ of \citetalias{SVR} gives us explicit polynomial bounds on the $i^{th}$ entropy number of RKHS generated by  $K^{\chi^2-\text{RBF}}_{\rho}$, and hence consistency follows.

\section{Assumptions for RFE in general function spaces}\label{sec:disc_assump} We now move beyond the premise that the functional space $\mathcal{F}$ admits the nestedness or denseness property. In this section, we will study closely the scope for risk-RFE under this generalized setup. 
\subsection{Assumptions}\label{sec:assump}
Consider the setting of risk minimization (regularized or non regularized) with respect to a given functional space $\mathcal{F}$. We first note the following assumptions:
\begin{enumerate}
\renewcommand{\theenumi}{(A\arabic{enumi})}
	\item\label{as:a1} Let $J$ be a subset of $\{1,\ldots,d\}$. Let $f_{P,\mathcal{F}^{J}}$ be the function that minimizes risk within $\mathcal{F}^{J}$ with respect to the measure $P$ on $\mathcal{X}\times\mathcal{Y}$. Define $\mathcal{F}^{\emptyset}=\mathcal{F}$. We assume that there exists a  $J_{\ast}$, such that, $\left|J_{\ast}\right|=d-d_{0}$ (where $d_{0}$ is the number of signals in the model) with $d\geq d_{0} > 0$. Then for any pair $(d_{1},d_{2})$ satisfying $d_{1}\leq d_{2}\leq d-d_{0}$, $\exists$ $J_{d_{1}}$ and $J_{d_{2}}$ with $J_{d_{1}}\subseteq J_{d_{2}}\subseteq J_{\ast}$ and $\left|J_{d_{1}}\right|=d_{1}$ and $\left|J_{d_{2}}\right|=d_{2}$, such that $\mathcal{R}^{\ast}_{L,P,\mathcal{F}^{J_{\ast}}} = \mathcal{R}^{\ast}_{L,P,\mathcal{F}^{J_{d_{1}}}} = \mathcal{R}^{\ast}_{L,P,\mathcal{F}^{J_{d_{2}}}}$.
\end{enumerate}	

\begin{rem}
Assumption \ref{as:a1} says that there exists a `path' from the original input space $\mathcal{X}$ to the correct lower dimensional space $\mathcal{X}^{J_{\ast}}$ in the sense of equality of minimized risk within $\mathcal{F}^{J}$s along this `path', and that set $J_{\ast}$ is maximal in terms of this property. Note that $\mathcal{J}$ need not be unique, and there can be more than one path going down to a given $J_{\ast}$. For simplicity, we assume that $J_{\ast}$ is unique,
leading to $\mathcal{X}^{J_{\ast}}$. 
\end{rem}

The following examples aim to show that assumption \ref{as:a1} above is in fact necessary in order for a well-defined recursive feature elimination algorithm to work under this setup.

\subsection{Necessity for existence of a path in \ref{as:a1}}
\begin{example}\label{example:existence}
Consider the following empirical risk minimization framework. Let $X=[-1,1]^{2}$ and let $Y=0$. Let $X_{1} \sim \mathcal{U}$ where $\mathcal{U}$ is some distribution on $[-1,1]$ and $X_{2}\equiv -X_{1}$. Let $\mathcal{F}$ be given as $\{c_1 X_{1}+c_2 X_{2},\: c_1,\:c_2 >1\}$, and then note that $\mathcal{F}$ neither admits the nested property nor is dense in $\mathcal{L}_{\infty}(\mathcal{X})$ or in any other space admitting the nested property. Let the loss function be the squared error loss, i.e., $L(x,y,f(x))=(y-f(x))^{2}$. By Definition~\ref{defn:fj} in section \ref{sec:feat_algo}, $\mathcal{F}^{\{1\}}=\{c_2 X_{2},\: c_2>0\}$ and $\mathcal{F}^{\{2\}}=\{c_1 X_{1},\: c_1>0\}$ and $\mathcal{F}^{\{1,2\}}=\{0\}$. We see that $\mathcal{R}_{L,P}(f_{P,\mathcal{F}})=\mathcal{R}_{L,P}(f_{P,\mathcal{F}^{\{1,2\}}})=0$ but both $\mathcal{R}_{L,P}(f_{P,\mathcal{F}^{\{1\}}})$ and $\mathcal{R}_{L,P}(f_{P,\mathcal{F}^{\{2\}}})$ are strictly positive. Hence even if the correct lower-dimensional functional space may have minimized risk same as that of the original functional space, if there does not exist a path going down to that space, the recursive algorithm will not work. Note that the minimizer of the risk belongs to $\mathcal F^{\{1,2\}}$, but there is no path from $\mathcal F$ to $\mathcal F^{\{1,2\}}$ in the sense of \ref{as:a1}.
\end{example}

\subsection{Necessity for equality in \ref{as:a1}} The following example shows that equality in \ref{as:a1} cannot be replaced by `$\leq$'.
\begin{example}\label{example:equality}
Consider another empirical risk minimization framework. Let $Y \sim U(-1,1)$ and $X\subset \mathbb{R}^{3}$ such that $Y=X_{3}=X_{2}+1=X_{1}-1$. Let $\mathcal{F}=\{c_{1}X_{1}+c_{2}X_{2}+c_{3}X_{3},\: c_{1},\:c_{2},\:c_{3}\geq 1\}$, and let the loss function be squared error loss. Now by definition, $\mathcal{F}^{\{1\}}=\{c_{2}X_{2}+c_{3}X_{3},\: c_{2},\:c_{3} \geq 1\}$, $\mathcal{F}^{\{2\}}=\{c_{1}X_{1}+c_{3}X_{3},\: c_{1},\:c_{3} \geq 1\}$, $\mathcal{F}^{\{3\}}=\{c_{2}X_{2}+c_{1}X_{1},\: c_{1},\:c_{2} \geq 1\}$, $\mathcal{F}^{\{1,2\}}=\{c_{3}X_{3},\: c_{3} \geq 1\}$, $\mathcal{F}^{\{1,3\}}=\{c_{2}X_{2},\: c_{2} \geq 1\}$, $\mathcal{F}^{\{2,3\}}=\{c_{1}X_{1},\: c_{1} \geq 1\}$, and $\mathcal{F}^{\{1,2,3\}}=\{0\}$. By simple calculations, we see that $\mathcal{R}_{L,P,\mathcal{F}}^{\ast}=\mathcal{R}_{L,P,\mathcal{F}^{\{1\}}}^{\ast}=\mathcal{R}_{L,P,\mathcal{F}^{\{2\}}}^{\ast}=4/3$, $\mathcal{R}_{L,P,\mathcal{F}^{\{3\}}}^{\ast}$ $=\mathcal{R}_{L,P,\mathcal{F}^{\{1,2,3\}}}^{\ast}=1/3$, $\mathcal{R}_{L,P,\mathcal{F}^{\{1,3\}}}^{\ast}=\mathcal{R}_{L,P,\mathcal{F}^{\{2,3\}}}^{\ast}=1$ and $\mathcal{R}_{L,P,\mathcal{F}^{\{1,2\}}}^{\ast}=0$.
Note that the correct dimensional subspace of the input space is $X^{\{1,2\}}$ and there exists paths leading to this space via $X\rightarrow X^{\{1\}}\rightarrow X^{\{1,2\}}$ (since $\mathcal{R}_{L,P,\mathcal{F}}^{\ast}=\mathcal{R}_{L,P,\mathcal{F}^{\{1\}}}^{\ast}> \mathcal{R}_{L,P,\mathcal{F}^{\{1,2\}}}^{\ast}$) or via $X\rightarrow X^{\{2\}}\rightarrow X^{\{1,2\}}$ (since $\mathcal{R}_{L,P,\mathcal{F}}^{\ast}=\mathcal{R}_{L,P,\mathcal{F}^{\{2\}}}^{\ast}> \mathcal{R}_{L,P,\mathcal{F}^{\{1,2\}}}^{\ast}$) in the sense of modified Assumption (A1). But there actually exists the blind path $X\rightarrow X^{\{3\}}$ too (since $\mathcal{R}_{L,P,\mathcal{F}}^{\ast}>\mathcal{R}_{L,P,\mathcal{F}^{\{3\}}}^{\ast}$), which does not lead to the correct subspace. Hence the recursive search in this case may not guarantee to lead to the correct subspace.
\end{example}

\subsection{Gap in the finite design setting} Define a feature as a \emph{signal} if and only if risk of the model gets inflated in its absence, and hence equivalently, if a feature does not contribute to the model at all, the increase in risk (regularized or non-regularized) on its removal should be inconsequential.  Then note that when the design size $d$ is finite, assumption \ref{as:a1} implies the following:
\begin{enumerate}
\renewcommand{\theenumi}{(A\arabic{enumi})}
\setcounter{enumi}{1}
	\item\label{as:a2} Let $\mathcal{J}_{1},\mathcal{J}_{2},\dots,\mathcal{J}_{N}$ be the exhaustive list of such paths from $\mathcal{X}$ to $\mathcal{X}^{J_{\ast}}$, and let $\widetilde{\mathcal{J}}:=\displaystyle\bigcup_{i=1}^{N}\mathcal{J}_{i}$. There exists $\epsilon_{0} > 0$ such that whenever $J \notin \widetilde{\mathcal{J}}$, $\mathcal{R}^{\ast}_{L,P,\mathcal{F}^{J}} \geq \mathcal{R}^{\ast}_{L,P,\mathcal{F}^{J_{\ast}}} + \epsilon_{0}$.
\end{enumerate}
	
Note also that equality in \ref{as:a1} guarantees that the recursive search will never select an important dimension $j \in J_{\ast}$ for redundancy because then \ref{as:a2} would be violated. Hence the equality in \ref{as:a1} will ensure that we follow the correct path recursively and \ref{as:a2} gives us a stopping condition to halt at the correct input space $\mathcal{X}^{J_{\ast}}$. Also note from discussions in section \ref{sec:nullmodel}, \ref{as:a1} and \ref{as:a2} are satisfied for nested or dense models. In the following section we will show that these are sufficient for a recursive feature elimination algorithm like risk-RFE to work in this generalized framework (in terms of consistency). 

\section{Theoretical results}\label{sec:theory} Note that Theorem \ref{thm:main2} in Section \ref{sec:consis} was stated under Condition \ref{cond:1}, but the result continues to hold if it is replaced by the more general Condition \ref{cond:2}.
\begin{cond}\label{cond:2}\ \\
Assumption \ref{as:a1} holds. 
\end{cond}
Then in the finite design setting we can assume that \ref{as:a2} holds as well. Now, we want to prove the main result under the more general premise of Condition \ref{cond:2}. Before that however, we will note a few relevant results that will help us in establishing this theorem.

\subsection{Additional results} We start off with the following lemma:

\begin{lem}\label{lem:rd-rp}
Let $(\mathcal{F},\|\cdot\|_{\mathcal{F}})$ be a separable functional space, such that the metric $\|\cdot\|_{\mathcal{F}}$ dominates pointwise convergence. Also we assume $\sup\|f\|_{\mathcal{F}}\leq C$ for some $C<\infty$ for all $f\in\mathcal{F}$. Let $L$ be a convex, locally Lipschitz loss function such that $L(x,y,f(x))\leq B$ for some $B<\infty$ for all $f\in\mathcal{F}$. Also assume that for fixed $n\geq1$, $\exists$ constants $a\geq1$ and $p\in (0,1)$ such that $\mathbb{E}_{D_{\mathcal{X}}\sim P^{n}_{\mathcal{X}}}e_{i}\left(\mathcal{F}, L_{\infty}(D_{\mathcal{X}})\right) \leq ai^{-\frac{1}{2p}},\quad i \geq 1$. Then, we have with probability greater than or equal to $ 1-e^{-\tau}$,
\begin{eqnarray*}
\lefteqn{\displaystyle\sup_{f\in\mathcal{F}}|\mathcal{R}_{L,P}(f)-\mathcal{R}_{L,D}(f)|\leq 2B\sqrt{\frac{2\tau}{n}}+\frac{10B\tau}{3n}}&&\\ &+& 4\max\left\{C_{1}(p)c_{L}(C)^{p}a^{p}B^{1-p}n^{-\frac{1}{2}},C_{2}(p)c_{L}(C)^{\frac{2p}{1+p}}a^{\frac{2p}{1+p}}B^{\frac{1-p}{1+p}}n^{-\frac{1}{1+p}}\right\}.
\end{eqnarray*}
for constants $C_1(p)$, $C_2(p)$ depending only on $p$.
\end{lem}

See Appendix \ref{sec:rd-rp_proof} for a proof. This lemma gives us a bound for the difference between the empirical risk of a function $f\in\mathcal{F}$ and its omniscient oracle risk. We now assume the premise of Condition \ref{cond:2}. The above lemma helps set up the next proposition, which aims to bound the difference in the regularized empirical risk of the empirical KM solutions obtained from spaces lying in the pathway hypothesized in Assumption~\ref{as:a1} in section \ref{sec:disc_assump}. 

\begin{prop}\label{prop:svm} 
Again we assume $P$ to be a probability measure on $\mathcal{X}\times\mathcal{Y}$, and that the input space $\mathcal{X}$ is a valid metric space. We will assume $L:\mathcal{X}\times\mathcal{Y}\times\mathbb{R} \mapsto [0,\infty]$ to be convex and locally Lipschitz continuous, satisfying $L(x,y,0) \leq 1$ for all $(x,y) \in \mathcal{X}\times\mathcal{Y}$. Again we assume $H$ to be the separable RKHS of a measurable kernel $k$ on $\mathcal{X}$ with $\|k\|_{\infty}\leq 1$, and that for fixed $n\geq1$, $\exists$ constants $a\geq1$ and $p\in (0,1)$ such that $\mathbb{E}_{D_{\mathcal{X}}\sim P^{n}_{\mathcal{X}}}e_{i}\left(id:H\mapsto L_{\infty}(D_{\mathcal{X}})\right) \leq ai^{-\frac{1}{2p}},\: i \geq 1$.
Now for a fixed $\lambda >0$, $\epsilon > 0$, $\tau > 0$, and $n \geq 1$, and for $J_{1},\:J_{2} \in\widetilde{\mathcal{J}}$ such that $J_{1} \subseteq J_{2} \subseteq J_{\ast}$, we have with probability $P^{n}$ not less than $1-2e^{-\tau}$,
\begin{align}
&\left|\mathcal{R}^{\text{reg},\lambda}_{L,D,H^{J_{2}}}\left(f_{D,\lambda,H^{J_{2}}}\right) - \mathcal{R}^{\text{reg},\lambda}_{L,D,H^{J_{1}}}\left(f_{D,\lambda,H^{J_{1}}}\right)\right|\\
&< A_{2}^{J_{1}}(\lambda) + A_{2}^{J_{2}}(\lambda) + 12B\sqrt{\frac{2\tau}{n}}+20B\frac{\tau}{n}+24K_{2}B^{1-p}\left(\frac{a^{2p}}{\lambda^{p}n}\right)^{\frac{1}{2}},\nonumber
\end{align}
where $A_{2}^{J_{1}}(\lambda)$ and $A_{2}^{J_{2}}(\lambda)$ are the approximation errors for the
two separate RKHS classes $H^{J_{1}}$ and $H^{J_{2}}$, $B:=c_{L}(\lambda^{-1/2})\lambda^{-1/2}+1$, and $K_{2}:=\max\left\{B^{p}/4,C_{1}(p)c_{L}(\lambda^{-\frac{1}{2}})^{p},C_{2}(p)c_{L}(\lambda^{-\frac{1}{2}})^{\frac{2p}{1+p}}\right\}$ is a constant depending only on $B$, $p$ and the Lipschitz constant $c_{L}(\lambda^{-1/2})$.
\end{prop}
See Appendix \ref{sec:prop_svm_proof} for a detailed proof of Proposition \ref{prop:svm}. 

Note that since $B\geq 1$ and $K_{2}\geq B^{p}/4$, we have that if $a^{2p}>\lambda^{p}n$,
\begin{align}
&\left|\mathcal{R}^{\text{reg},\lambda}_{L,D,H^J}\left(f_{D,\lambda,H^{J}}\right) - \mathcal{R}_{L,P,H^{J}}^{\ast}\right|\nonumber\\
&\leq \mathcal{R}_{L,D}(0)+\mathcal{R}_{L,P}(0)\leq 2 < 3B \leq 12K_{2}B^{1-p}\left(\frac{a^{2p}}{\lambda^{p}n}\right)^{\frac{1}{2}}.
\end{align}

Similarly, since $B\geq 1$ and $K_{2}\geq B^{p}/4$, we have for $a^{2p}>\lambda^{p}n$,
\begin{align*}
&\mathcal{R}^{\text{reg},\lambda}_{L,P,H^J}\left(f_{D,\lambda,H^{J}}\right)-\mathcal{R}^{\ast}_{L,P,H^{J}}\nonumber\\
 &\leq \lambda\left\|f_{D,\lambda,H^{J}}\right\|^{2}_{H^{J}} +\mathcal{R}_{L,D}\left(f_{D,\lambda,H^{J}}\right)+\mathcal{R}_{L,P}\left(f_{D,\lambda,H^{J}}\right)\nonumber\\&\leq \mathcal{R}_{L,P}(0)+\mathcal{R}_{L,P}\left(f_{D,\lambda,H^{J}}\right) \leq 1+B \leq 2B \leq 8K_{2}B^{1-p}\left(\frac{a^{2p}}{\lambda^{p}n}\right)^{\frac{1}{2}}.
\end{align*}
Consequently we obtain the following two corollaries:
\begin{cor}\label{cor:svm} Assume the conditions of Proposition \ref{prop:svm}. For any $J$ and all $\epsilon > 0$, $\tau > 0$, and $n \geq 1$, we have with $P^{n}$ probability $> 1 - e^{-\tau}$,
\begin{align*}
&\left|\mathcal{R}^{\text{reg},\lambda}_{L,D,H^{J}}\left(f_{D,\lambda,H^{J}}\right) - \mathcal{R}_{L,P,H^{J}}^{\ast}\right| \\
<& A_{2}^{J}(\lambda) + 6B\sqrt{\frac{2\tau}{n}}+10B\frac{\tau}{n}+12K_{2}B^{1-p}\left(\frac{a^{2p}}{\lambda^{p}n}\right)^{\frac{1}{2}},
\end{align*}
where $K_{2}$ is as before. Additionally, if $J \in\widetilde{\mathcal{J}}$, we can replace $\mathcal{R}_{L,P,\mathcal{F}^{J}}^{\ast}$ in the above inequality by $\mathcal{R}_{L,P,\mathcal{F}}^{\ast}$.
\end{cor}

\begin{cor}\label{cor:oracle_svm} {\textsc{Oracle Inequality for KM:}} Assume the conditions of Proposition \ref{prop:svm}. For any $J$ and all $\epsilon > 0$, $\tau > 0$, and $n \geq 1$, we have with $P^{n}$ probability $> 1 - e^{-\tau}$,
\begin{align*}
&\mathcal{R}^{\text{reg},\lambda}_{L,P,H^J}\left(f_{D,\lambda,H^{J}}\right) - \mathcal{R}_{L,P,H^{J}}^{\ast} \\
&< A_{2}^{J}(\lambda)+4B\sqrt{\frac{2\tau}{n}}+\frac{20B\tau}{3n} + 8K_{2}B^{1-p}\left(\frac{a^{2p}}{\lambda^{p}n}\right)^{\frac{1}{2}},
\end{align*}
where $K_{2}$ is as before.
\end{cor}
Proposition \ref{prop:svm} and Corollaries \ref{cor:svm} and \ref{cor:oracle_svm} jointly imply Lemma \ref{lem:aims2}, given below.
We will now assume the rest of the conditions stated before Theorem \ref{thm:main2}. We assume that the regularization constant $\lambda_{n}$ converges to $0$ and that $\lim_{n \rightarrow \infty}\lambda_n n = \infty$. To establish explicit rates for our algorithm, we further assume that the approximation error $A^{J}_{2}(\lambda)$ assumes a polynomial bound.

\begin{lem}\label{lem:aims2} Assume the conditions of Theorem \ref{thm:main2} in section \ref{sec:consis}. Then for a sequence $\tau = o\left(n^{\frac{2\beta}{2\beta+1}}\right)$, the following statements hold:
\begin{enumerate}
\renewcommand{\theenumi}{\roman{enumi}}
\item\label{itm:aim1_svm} For $J_{1},\:J_{2} \in \widetilde{\mathcal{J}}$ such that $J_{1}\subseteq J_{2}\subseteq J_{\ast}$, there is a positive sequence $\{\epsilon_{n}\}$ with $\epsilon_n \rightarrow 0$ for which we have with $P^n$ probability greater than $1-2e^{-\tau}$,
\begin{align*}
\mathcal{R}^{\text{reg},\lambda_n}_{L,D,H^{J_{2}}}\left(f_{D,\lambda_{n},H^{J_{2}}}\right) \leq \mathcal{R}^{\text{reg},\lambda_n}_{L,D,H^{J_{1}}}\left(f_{D,\lambda_{n},H^{J_{1}}}\right) + \epsilon_{n}.
\end{align*}
\item\label{itm:aim2_svm} For $J_{1} \in\widetilde{\mathcal{J}}$ and $J_{2} \notin \widetilde{\mathcal{J}}$ and for $J_{1} \subset J_{2}$, there is a positive sequence $\{\epsilon_{n}\}$ with $\epsilon_n \rightarrow 0$, for which we have with $P^n$ probability greater than $1-2e^{-\tau}$,
\begin{align*}
 \mathcal{R}^{\text{reg},\lambda_n}_{L,D}\left(f_{D,\lambda_{n},H^{J_{2}}}\right) \geq \mathcal{R}^{\text{reg},\lambda_n}_{L,D}\left(f_{D,\lambda_{n},H^{J_{1}}}\right) + \epsilon_{0} -\epsilon_{n}.
\end{align*}
\item\label{itm:aim3_svm} \textsc{Oracle Property for risk-RFE in KM:} The infinite-sampled regularized risk for the empirical solution $f_{D,\lambda_{n},H^{J}}$, $\mathcal{R}^{\text{reg},\lambda_n}_{L,P,H^J}\left(f_{D,\lambda_{n},H^{J}}\right)$ converges in measure to $\mathcal{R}_{L,P,H}^{\ast}$ (and hence to $\mathcal{R}_{L,P}^{\ast}$ if the RKHS $H$ is dense in $\mathcal{L}_{\infty}\left(\mathcal{X}\right)$) iff $J \in \widetilde{\mathcal{J}}$.
\end{enumerate}
\end{lem}

The proof of Lemma \ref{lem:aims2} is given in Appendix \ref{sec:aims2_proof}. 
We are now ready to prove Theorem \ref{thm:main2} under condition \ref{cond:2}. 

\subsection{Proof of Theorem \ref{thm:main2} (from section \ref{sec:consis})}\label{sec:main2_proof}

\begin{proof} \eqref{itm:part1_svr} Let $\mathcal{X}^{J_{\ast}}$ be the correct input space and $J_{\ast}$ be the correct set of dimensions to be removed with $|J_{\ast}|=d-d_{0}$. To prove the first part of Theorem \ref{thm:main2}, we show that, starting with the input space $\mathcal{X}$, the probability that we reach the space $\mathcal{X}^{J_{\ast}}$ is $1$ asymptotically. First let us assume that there exists only one correct `path' from $\mathcal{X}$ to $\mathcal{X}^{J_{\ast}}$. Let $\mathcal{J}^{\circ}$ be that correct path and $\mathcal{J}^{\circ}=\{J^{\circ}_{0}\equiv\{\cdot\}, J^{\circ}_{1},\dots,J^{\circ}_{d-d_{0}}\equiv J_{\ast}\}$. 

For notational ease, let us further define,
\begin{align}\label{eq:shorts}
\nonumber
&\mathcal{R}\mathcal{R}^{Q_1}_{Q_2}\left(J_{1},J_{2}\right):= \mathcal{R}^{\text{reg},\lambda_n}_{L,Q_2,H^{J_{1}}}\left(f_{Q_1,\lambda_{n},H^{J_{1}}}\right) - \mathcal{R}^{\text{reg},\lambda_n}_{L,Q_2,H^{J_{2}}}\left(f_{Q_1,\lambda_{n},H^{J_{2}}}\right)\nonumber\\
&\mathcal{R}\mathcal{R}^{Q_1}_{Q_2}\left(J\right):=\mathcal{R}^{\text{reg},\lambda_n}_{L,Q_2,H^{J}}\left(f_{Q_1,\lambda_{n},H^{J}}\right) - \mathcal{R}^{\ast}_{L,P,H}
\end{align}

From the proof of \eqref{itm:aim1_svm} in Appendix \ref{sec:aims2_proof}, we have $\mathcal{R}\mathcal{R}^D_D\left(J^{\circ}_{i+1},J^{\circ}_i\right) \leq \epsilon_{n}$ with probability at least $1-2e^{-\tau}$ for $\epsilon_{n}=(2c+24\sqrt{2\tau}+48K_2a^{2p})n^{-\frac{\beta}{2\beta+1}} + 40\tau n^{-\frac{4\beta+1}{2(2\beta+1)}}$. Now letting $J_{i+1} \neq J^{\circ}_{i+1}$ be any other $J$ such that $J^{\circ}_{i}\subset J_{i+1}$ with $|J_{i+1}|=|J^{\circ}_{i}|+1$, we have from \eqref{eq:hj1-hj2} in Appendix \ref{sec:aims2_proof} that $\mathcal{R}\mathcal{R}^D_D\left(J_{i+1},J_i\right) > \epsilon_{0} - \epsilon_{n}$ with probability at least $1-2e^{-\tau}$. Now if we choose $\tau=o(n^{\frac{2\beta}{2\beta+1}})$ with $\tau\rightarrow\infty$, then we see $\epsilon_n=O(n^{-\frac{\beta}{2\beta+1}})$, and hence $\delta_{n}\equiv\epsilon_0-\epsilon_{n}$ satisfies the second inequality with the condition that $\delta_{n}=\epsilon_0-O(n^{-\frac{\beta}{2\beta+1}})$ with $\delta_{n}\rightarrow 0$. Now since $\epsilon_{0}$ is a fixed constant, $\exists N_{\epsilon_{0}}>0$ such that $\forall n\geq N_{\epsilon_{0}}$, $2\epsilon_{n}\leq \epsilon_{0}$. Without loss of generality we assume that $n\geq N_{\epsilon_{0}}$. Then we have the condition that $\mathcal{R}\mathcal{R}^D_D\left(J^{\circ}_{i+1},J^{\circ}_i\right) \leq \delta_{n}$
with probability at least $1-2e^{-\tau}$.

Then, 
\begin{eqnarray*}
&&P\left(\text{\emph{`risk-RFE finds the correct space'}}\right)\\
&\geq& P\left(\text{\emph{`risk-RFE follows the path $\mathcal{J}^{\circ}$ to the correct dimension space'}}\right)\\
&=& P\left(J_{0}:=J^{\circ}_{0},\:J_{1}:=J^{\circ}_{1},\dots,J_{d-d_{0}}:=J^{\circ}_{d-d_{0}},J_{d-d_{0}+1}:=\emptyset \right)\\
&=& P\left(J_{0}:=J^{\circ}_{0}\right)P\left(J_{1}:=J^{\circ}_{1}\big|J^{\circ}_{0}\right)\cdots P\left(J_{d-d_{0}}:=J^{\circ}_{d-d_{0}}\big|J^{\circ}_{0},\dots,J^{\circ}_{d-d_{0}-1}\right)\\
&&P\left(J_{d-d_{0}+1}:=\emptyset\big|J^{\circ}_{0},\dots,J^{\circ}_{d-d_{0}} \right),
\end{eqnarray*}
where `$J_{d-d_{0}+1}:=\emptyset$' means the algorithm stops at that step. Note that $P\left(J_{0}:=J^{\circ}_{0}\right)=1$ and then observe,
\begin{align*}
&P\left(J_{i+1}:=J^{\circ}_{i+1}\big|J^{\circ}_{0},\dots,J^{\circ}_{i}\right)\\
=&P\left(J_{i+1}:=J^{\circ}_{i+1}\big|J^{\circ}_{i}\right)\quad(\:\because \: \{J^{\circ}_{0},\dots,J^{\circ}_{i-1}\}\:\text{\emph{have already been removed}})\\
=&P\Bigl( \mathcal{R}\mathcal{R}_D^D\left(J^{\circ}_{i+1},J^{\circ}_{i}\right) \leq \delta_{n}\:,\: \mathcal{R}\mathcal{R}_D^D\left(J^{\circ}_{i+1},J^{\circ}_{i}\right) < \mathcal{R}\mathcal{R}_D^D\left(J^{\bullet}_{i+1},J^{\circ}_{i}\right) \:\:\forall J^{\bullet}_{i+1}\neq J^{\circ}_{i+1}\Bigr)\\
\geq & P\Bigl(\mathcal{R}\mathcal{R}_D^D\left(J^{\circ}_{i+1},J^{\circ}_{i}\right) \leq \delta_{n}\:,\: \delta_{n} < \mathcal{R}\mathcal{R}_D^D\left(J^{\bullet}_{i+1},J^{\circ}_{i}\right)\:\:\forall J^{\bullet}_{i+1}\neq J^{\circ}_{i+1}\Bigr)\\
\geq & 1 - P\left( \mathcal{R}\mathcal{R}_D^D\left(J^{\circ}_{i+1},J^{\circ}_{i}\right) > \delta_{n}\right) - \displaystyle\sum_{J^{\bullet}_{i+1}\neq J^{\circ}_{i+1}}P\left( \mathcal{R}\mathcal{R}_D^D\left(J^{\bullet}_{i+1},J^{\circ}_{i}\right)\leq \delta_{n}\right)\\
\geq & 1 - 2e^{-\tau} - 2(d-i-1)e^{-\tau}\:=\: 1 - 2(d-i)e^{-\tau}.
\end{align*}
Also note that,
\begin{align*}
&P\left(J_{d-d_{0}+1}:=\emptyset\big|J^{\circ}_{0},\dots,J^{\circ}_{d-d_{0}} \right)\\ = & P\left( \mathcal{R}\mathcal{R}_D^D\left(J_{d-d_{0}+1},J^{\circ}_{d-d_{0}}\right)>\delta_{n}\:\:\forall J_{d-d_{0}+1} \supseteq J^{\circ}_{d-d_0}\right) \geq 1 - 2d_{0}e^{-\tau}.
\end{align*}
Hence $P\left(\text{\emph{`risk-RFE finds the correct space'}}\right) \geq \displaystyle\prod_{i=0}^{d-d_{0}}\left(1 - 2(d-i)e^{-\tau}\right)$.
Now for $\tau=o(n^{\frac{2\beta}{2\beta+1}})$ with $\tau\rightarrow\infty$, $P\left(\text{\emph{`risk-RFE finds the correct space'}}\right) \rightarrow \: 1$ as $n\rightarrow \infty$.

Now let us prove the same assertion for the case when there is more than one correct `path' from $\mathcal{X}$ to $\mathcal{X}^{J_{\ast}}$. Let $\mathcal{J}_{1},\dots,\mathcal{J}_{N}$ be an enumeration of all possible such paths. Define `C-set' for a given set $J_{i}$ (where index $i$ denotes the $i^{th}$ iteration of RFE) as
$CS(J_{i}):=\{J_{i+1}\::\:J_{i},\: J_{i+1} \in\mathcal{J}_{k}\:\text{\emph{for some}}\:k\}$. Now,
\begin{eqnarray*}
&&P\left(\text{\emph{`risk-RFE finds the correct space'}}\right)\\
&\geq& P\left(J_{0}:=J^{\circ}_{0},\:J_{1}:=J^{\circ}_{1}\in CS(J^{\circ}_{0}),\dots,J_{d-d_{0}+1}:=\emptyset \right)\\
&=& P\left(J_{0}:=J^{\circ}_{0}\right)P\left(J_{1}:=J^{\circ}_{1}\in CS(J^{\circ}_{0})\big|J^{\circ}_{0}\right)\cdots P\left(J_{d-d_{0}+1}:=\emptyset\big|J^{\circ}_{d-d_{0}} \right).
\end{eqnarray*}
Again as before $P\left(J_{0}:=J^{\circ}_{0}\right)=1$ and $P\left(J_{d-d_{0}+1}:=\emptyset\big|J^{\circ}_{d-d_{0}} \right)\geq 1 - 2d_{0}e^{-\tau}$. Now note,
\begin{align*}
&P\left(J_{i+1}:=J^{\circ}_{i+1}\in CS\left(J^{\circ}_{i}\right)\big|J^{\circ}_{i}\right)\\
&\geq P\left(\mathcal{R}\mathcal{R}_D^D\left(J^{\circ}_{i+1},J^{\circ}_{i}\right)\leq \delta_{n}\:\:\forall J^{\circ}_{i+1}\in CS\left(J^{\circ}_{i}\right) \right.,\\
&\quad \quad \quad \left. \delta_{n} < \mathcal{R}\mathcal{R}_D^D\left(J^{\bullet}_{i+1},J^{\circ}_{i}\right)\:\:\forall J^{\bullet}_{i+1}\notin CS\left(J^{\circ}_{i}\right)\right)\\
&\geq 1 - \sum_{J^{\circ}_{i+1}\in CS\left(J^{\circ}_{i}\right)}P\left(\mathcal{R}\mathcal{R}_D^D\left(J^{\circ}_{i+1},J^{\circ}_{i}\right) > \delta_{n}\right)\\
&\quad\quad - \sum_{J^{\bullet}_{i+1}\notin CS\left(J^{\circ}_{i}\right)}P\left(\mathcal{R}\mathcal{R}_D^D\left(J^{\bullet}_{i+1},J^{\circ}_{i}\right) \leq \delta_{n}\right)\\
&\geq 1 - 2\left|CS\left(J^{\circ}_{i}\right)\right|e^{-\tau} - 2\left|CS\left(J^{\circ}_{i}\right)^{c}\right|e^{-\tau}\:=\: 1 - 2(d-i)e^{-\tau},
\end{align*}
since $\left|CS\left(J^{\circ}_{i}\right)\right| + \left|CS\left(J^{\circ}_{i}\right)^{c}\right|=d-i$. Hence again we have that,\\ $P\left(\text{\emph{`risk-RFE finds the correct space'}}\right) \geq \displaystyle\prod_{i=0}^{d-d_{0}}\left(1 - 2(d-i)e^{-\tau}\right)$. Now for $\tau=o(n^{\frac{2\beta}{2\beta+1}})$ with $\tau\rightarrow\infty$, $P\left(\text{\emph{`risk-RFE finds the correct space'}}\right) \rightarrow \: 1$ as $n\rightarrow \infty$.

\eqref{itm:part2_svr} To prove the second part of Theorem \ref{thm:main2} just observe that if $J_{\text{end}}$ is the last iteration of the algorithm in risk-RFE, then from~\eqref{eq:oracle_eq_svm} in Appendix \ref{sec:aims2_proof}, and recycling arguments given at the beginning of the first part of the proof we have that
\begin{align*}
&P\left(\left|\mathcal{R}\mathcal{R}_D^D\left(J_{\text{end}}\right)\right|\leq \delta_{n}\right)\\
=& P\left(\left|\mathcal{R}\mathcal{R}_D^D\left(J_{\ast}\right)\right|\leq \delta_{n}\right)P\left(J_{\text{end}}=J_{\ast}\right)\\
&\quad \quad \quad + P\left(\left|\mathcal{R}\mathcal{R}_D^D\left(J_{\text{end}}\right)\right|\leq \delta_{n}\big|J_{\text{end}}\neq J_{\ast}\right)P\left(J_{\text{end}}\neq J_{\ast}\right)\\
\geq & P\left(\left|\mathcal{R}\mathcal{R}_D^D\left(J_{\ast}\right)\right|\leq \delta_{n}\right)P\left(J_{\text{end}}=J_{\ast}\right)\\
\geq & (1-e^{-\tau})\displaystyle\prod_{i=0}^{d_{0}}\left(1 - 2(d-i)e^{-\tau}\right).
\end{align*}
So for $\tau=o(n^{\frac{2\beta}{2\beta+1}})$ with $\tau\rightarrow\infty$,\\ $P\left(\left| \mathcal{R}^{\text{reg},\lambda_n}_{L,D,H^{J_{\text{end}}}}\left(f_{D,\lambda_{n},H^{J_{\text{end}}}}\right) - \mathcal{R}^{\ast}_{L,P,H}\right|\leq \delta_{n}\right) \rightarrow \: 1$ with $n\rightarrow\infty$.
\end{proof}
\begin{rem}
Although~\eqref{eq:oracle_eq_svm} in Appendix \ref{sec:aims2_proof} was asserted for $\eta_{n}$, we do have $\eta_{n} < \epsilon_n < \delta_{n}$ $\forall n\geq N_{\epsilon_0}$, so the proof for the second part of the theorem will hold true for $\delta_{n}$.
\end{rem}

\section{Discussion on the high dimensional framework when dimension $d$ grows with sample size $n$}\label{sec:large_p} Our results till now have been derived under the assumption that $\dim(\mathcal{X})=d$ is fixed. In this section, we discuss the theoretical properties of risk-RFE when both $d$, the dimension of $\mathcal{X}$, and $d_0$, the number of relevant features, go to infinity with $n$. We provide modified arguments to achieve consistency as achieved in fixed design settings.

\subsection{The modified algorithm} Assume that $\mathcal{X}\in \mathbb{R}^d$, and that the observed data $D=\{(X_1,Y_1),\dots,(X_n, Y_n)\} \sim\:\text{i.i.d.}\:P^d_{\mathcal{X}\times \mathcal{Y}}$, where the probability distribution of the design depends on the dimension $d$ of the input space $\mathcal{X}$. 
The modified feature selection algorithm is given below. 

\begin{figure}
\begin{center}
\includegraphics[width=0.6\textwidth]{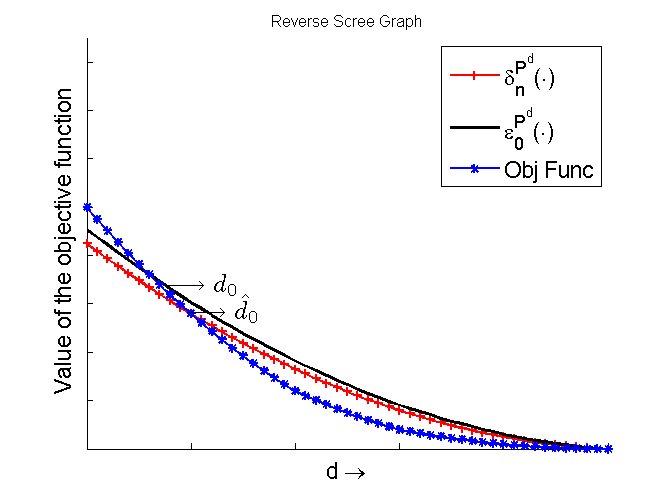}
\caption{Stopping rule for the modified algorithm in the growing design size setting: A potential case}
\label{fig:delta_pn}
\end{center}
\end{figure}

\begin{algo}\label{algo:svm_d} Replace the stopping condition in Algorithm \ref{algo:svm} in section \ref{sec:feat_algo} from $\delta_n$ to $\delta^{P^d}_n\left(d-|J|\right)$, where $\delta^{P^d}_n\left(\cdot\right)$ is a known positive function intrinsic to the design $P^d$.
\end{algo}
The only modification of the algorithm lies in the stopping rule. The fixed constant $\delta_n$ in the fixed design problem is replaced by the function $\delta^{P^d}_n: \{1,\dots,d\}\mapsto \mathbb{R}$. Figure \ref{fig:delta_pn} shows a visual representation of the stopping condition in this case. Extending the rationale from the fixed design case, $\delta^{P^d}_n$ acts as a functional upper bound for the difference of the regularized risks (that we use as the objective function for risk-RFE). In other words, at iteration $i$ of the algorithm, $\delta^{P^d}_n\left(d-i\right)$ acts as the maximal allowance that the difference of regularized risks (between models at subsequent iterations) can attain at this iteration. Hence, the point where this difference finally jumps above $\delta^{P^d}_n$, is where our algorithm is stopped, and the features left in the model are retained as potential signals. 

\subsection{Modified assumptions} To achieve consistency for this algorithm, we need to revisit our assumptions again. Let us consider the most general framework (Condition \ref{cond:2} in section \ref{sec:theory}). We keep assumption \ref{as:a1} (from section \ref{sec:disc_assump}) fixed, but \ref{as:a2} does not follow anymore since the fixed gap made sense only in a fixed design problem. In a varying design problem, this gap might diminish and shrink to $0$ as $d$ tends to infinity. Hence \ref{as:a2} is modified to \ref{as:a2*} to accommodate these dynamics and is given below:    

\begin{enumerate}
\renewcommand{\theenumi}{(A\arabic{enumi}*)}
\setcounter{enumi}{1}
	\item\label{as:a2*} Let $\mathcal{J}_{1},\mathcal{J}_{2},\dots$ be an enumeration of paths from $\mathcal{X}$ to $\mathcal{X}^{J_{\ast}}$, and let $\widetilde{\mathcal{J}}:=\displaystyle\bigcup_{i}\mathcal{J}_{i}$. There exists a monotonically decreasing function $\epsilon^{P^d}_{0}(\cdot)$, intrinsic to the problem and positive, such that for $J_1 \in \widetilde{\mathcal{J}},\: J_2 \notin \widetilde{\mathcal{J}}$ with $|J_2|=|J_1|+1$, we have 
\begin{align}\label{eq:a2_new}
\mathcal{R}^{\ast}_{L,P^d,\mathcal{F}^{J_2}} \geq \mathcal{R}^{\ast}_{L,P^d,\mathcal{F}^{J_1}} + \epsilon^{P^d}_{0}\left(d-|J_1|\right).
\end{align}
\end{enumerate}	
Our assumption now reflects the varying gap size. For a problem $P^d$, $\epsilon^{P^d}_0(\cdot)$ is a strictly positive and monotonically decreasing function from $\{1,\dots,d\}\mapsto \mathbb{R}$, such that $\epsilon^{P^d}_0(d-d_0)$ goes to zero in limit, when $d \rightarrow \infty$ ($d_0$ can potentially grow with $n$ as well, but we will restrict to the case when $d-d_0$ necessarily grows with $n$, for example when $d_0=O(d^{\alpha})$ for $0<\alpha <1$ and $d$ grows with $n$).  Hence there are two different asymptotic conditions acting on $\delta^{P^d}_n(\cdot)$ here, with $\delta^{P^d}_n(\cdot)\rightarrow \epsilon^{P^d}_{0}(\cdot)$ as $n\rightarrow \infty$, and additionally $\delta^{P^d}_n(d-d_0)\rightarrow 0$ as $d$ and $n$ go to infinity.

\subsection{Regularity conditions} With the growing design size, the regularity conditions need to be restated as well. The entropy bound for a RKHS $H$ (and the approximation bound on the regularization error) may very well depend on the growth of dimensions, and such relationships are characteristic of the kernel $k$ that we use, as well as the input-output space, and hence should be analyzed on a case by case basis. A generalized bound that governs the relationship is not only difficult to establish but may also be suboptimal in many other settings. We restrict our discussion to the setting of Condition \ref{cond:1} from Section \ref{sec:consis}. We also assume regularity conditions \ref{itm:rel} below:
\begin{enumerate}
\renewcommand{\theenumi}{(RC)}
\item There exist constants $\tilde{a}\geq1$ and some $p\in (0,1)$ such that $\mathbb{E}_{D_{\mathcal{X}}\sim P^{d,n}_{\mathcal{X}}}e_{i}\left(id:H\mapsto L_{\infty}(D_{\mathcal{X}})\right) \leq O\left(f(d)\right)  \tilde{a} i^{-\frac{1}{2p}},\: i \geq 1$. In addition there exists a $\tilde{c}>0$ and $\beta \in (0,1]$ such that $A_{2}^{J_{\ast}}(\lambda)\leq  O\left(g(d_0)\right) \tilde{c}\lambda^{\beta}$ for $J_{\ast}$ defined before and $\|J_{\ast}\|=d_0$, and for all $\lambda \geq 0$, for some functions $f(\cdot)$ and $g(\cdot)$ on $\mathbb{N}\mapsto \mathbb{R}$ \label{itm:rel}. We also assume that there exists a $\gamma \in \left(0,\left.\frac{\beta}{2\beta+1}\right.\right]$ such that,
\begin{itemize}
\item $O\left(g(d_0)\right) \leq C_1 n^{\frac{\beta}{2\beta+1}-\gamma}$,
\item $O\left(f(d)\right) \leq C_2 n^{\frac{\beta}{2\beta+1}-\gamma}$,
\item $d = o(e^{0.5n^{\frac{2\beta }{2\beta +1}}})$.
\end{itemize}

\end{enumerate}

\subsection{Consistency under \ref{itm:rel}}\label{sec:consis_d} Under condition \ref{itm:rel}, it can be seen that the modifications required for bounds in Lemma \ref{lem:rd-rp} -- Corollary \ref{cor:oracle_svm} from section \ref{sec:theory} can be achieved by replacing $a$ by $O\left(f(d)\right)  \tilde{a}$ and $c$ by $O\left(f(d_0)\right)  \tilde{c}$. Lemma \ref{lem:aims2} can now be restated by replacing $\epsilon_n$ by $\epsilon_{n,d}=(2O\left(f(d_0)\right)  \tilde{c}+24\sqrt{2\tau}+48K_2a^{2p} O\left(f(d)\right)^{2p} )n^{-\frac{\beta}{2\beta+1}} + 40\tau n^{-\frac{4\beta+1}{2(2\beta+1)}}$. Suppose now that we can show $\epsilon_{n,d}$ goes to $0$ in limit, Then Statements \eqref{itm:aim1_svm} and \eqref{itm:aim3_svm} in  Lemma \ref{lem:aims2} will continue to hold under slightly restated versions ($P^n$ is replaced with $P^{d,n}$ to denote the appropriate probability measure), while Statement \eqref{itm:aim2_svm} can be changed to the following:

\begin{enumerate}
\renewcommand{\theenumi}{\roman{enumi}*}
\setcounter{enumi}{1}
\item\label{itm:aim2_svm*} For $J_{1} \in\widetilde{\mathcal{J}}$ and $J_{2} \notin \widetilde{\mathcal{J}}$ and for $|J_{2}|= |J_{1}|+1$, there is a positive sequence $\{\epsilon_{n,d}\}$ with $\epsilon_{n,d} \rightarrow 0$, such that we have with $P^{d,n}$ probability greater than $1-2e^{-\tau}$,
\begin{align*}
 \mathcal{R}^{\text{reg},\lambda_n}_{L,D,H^{J_{2}}}\left(f_{D,\lambda_{n},H^{J_{2}}}\right) \geq \mathcal{R}^{\text{reg},\lambda_n}_{L,D,H^{J_{1}}}\left(f_{D,\lambda_{n},H^{J_{1}}}\right) + \epsilon^{P^d}_{0}\left(d-|J_1|\right) - \epsilon_{n,d}.
\end{align*}
\end{enumerate}

Under the modified statement, consistency can be established. It can be easily observed that the initial steps in the proof of Theorem \ref{thm:main2} in section \ref{sec:main2_proof} continue to hold by taking $\delta^{P^d}_n\left(d-\left\vert{J}\right\vert\right)=\epsilon^{P^d}_0\left(d-\left\vert{J}\right\vert\right)-\epsilon_{n,d}$ for design $\mathcal{X}^{J}$, and now we further assume that $\sup_{d\in \mathbb{N}}\underset{n\rightarrow \infty}{\lim\inf} \frac{\epsilon^{P^d}_0(d-d_0)}{\epsilon_{n,d}}>2$. This allows us to define a sequence $\{N_1,\dots,N_d,\dots\}$, such that $2\epsilon_{n,d}\leq \epsilon^{P^d}_0(d-d_0)$, whenever $n>N_d$. Since $\epsilon^{P^d}_0(\cdot)$ is a decreasing function, the subsequent steps follow and we arrive at,
\begin{align*}
P\left(\text{\emph{`RFE finds the correct space'}}\right) &\geq \displaystyle\prod_{i=0}^{d-d_{0}}\left(1 - 2(d-i)e^{-\tau}\right)\\
&\gtrsim \left(1 - 2de^{-\tau}\right)^d,
\end{align*}
where the last approximate inequality follows assuming $2de^{-\tau}<1$ for sufficiently large $n$, and $\tau=o(n^{\frac{2\beta}{2\beta+1}})$ with $\tau\rightarrow\infty$. Now for the limiting infinite product to converge to $1$ when $n,d\rightarrow \infty$, we have
\begin{align*}
\left(1 - 2de^{-\tau}\right)^d = \left(\left(1 - \frac{2d}{e^{\tau}}\right)^{-\frac{e^{\tau}}{2d}}\right)^{-\frac{2d^2}{e^{\tau}}}.
\end{align*}
Hence if we assume $d^2 e^{-\tau} \rightarrow 0$, the above quantity converges to $1$ in limit. Consequently for consistency results to hold, $d$ needs to grow slower than a certain rate in terms of the sample size $n$. Since $\tau$ can be chosen to be $o(n^{\frac{2\beta}{2\beta+1}})$ implies that $d e^{-\tau/2} \approx d e^{- 0.5n^{\frac{2\beta}{2\beta+1}}}$, and hence $d = o(e^{0.5n^{\frac{2\beta }{2\beta +1}}})$ suffices. 

We now need to ensure that asymptotically $\epsilon_{n,d}$ goes to $0$. Since $\tau=o(n^{\frac{2\beta}{2\beta+1}})$, this reduces $\epsilon_{n,d}$ to $\epsilon_{n,d}=O\left(f(d)\right) n^{-\frac{\beta}{2\beta+1}} + O\left(g(d_0)\right) n^{-\frac{\beta}{2\beta+1}}  + o(1)$. Fix a constant $\gamma \in \left(0,\left.\frac{\beta}{2\beta+1}\right.\right]$ such that $\epsilon_{n,d_n}=O(n^{-\gamma})$, then $O\left(g(d_0)\right) \leq C_1 n^{\frac{\beta}{2\beta+1}-\gamma}$ and $O\left(f(d)\right) \leq C_2 n^{\frac{\beta}{2\beta+1}-\gamma}$ must hold to ensure  $\epsilon_{n,d}$ goes to $0$.
\begin{thm}
Assume regularity conditions \ref{itm:rel} along with the rest of the conditions before Theorem \ref{thm:main2}. Then there exists $\delta^{P^d}_{n}(\cdot)=\epsilon^{P^d}_0-O(n^{-\gamma})$ where $\gamma \in \left(0,\left.\frac{\beta}{2\beta+1}\right.\right]$, such that statements \ref{itm:part1_svr} and \ref{itm:part2_svr} from Theorem \ref{thm:main2} in section \ref{sec:consis} continue to hold for risk-RFE algorithm with $\{\delta^{P^d}_{n}(\cdot), \lambda_n\}$.
\end{thm}

Let us now look at the allowed dimensionality growth under certain forms of $f(\cdot)$ and $g(\cdot)$.
\begin{enumerate}
\renewcommand{\theenumi}{(RC\arabic{enumi})}
\item\label{itm:rc1} $f(d)=c_1$ and $g(d_0)=c_2$. Under this setting we can allow rates as high as $d = o(e^{0.5n^{\frac{2\beta}{2\beta +1}}})$ with $d_0= O(d^{\alpha})$ with $0<\alpha<1$ and the algorithm will continue to learn at $\gamma =\frac{\beta}{2\beta+1}$.
\item\label{itm:rc2} $f(d)=e^d$ and $g(d_0)=e^{d_0}$. Under this setting it can be seen that we need $d =O(\log n)$. We can still allow $d_0= O(d^{\alpha})$ with $0<\alpha<1$, and the algorithm learns at $0<\gamma <\frac{\beta}{2\beta+1}$.
\item \label{itm:rc3} \textsc{Dimensionality growth in the Gaussian RBF kernel:} $f(d)=e^d$ and $g(d)=d_0^{c d_0}$. Under this, we can continue to have $d =O(\log n)$ but now we need $d_0 \log d_0 = O(\log n)$ and the algorithm learns at $0<\gamma <\frac{\beta}{2\beta+1}$.
\end{enumerate}
For a more detailed discussion on \ref{itm:rc3}, please see the \ref{sec:suppA}.
  
\section{Simulation studies}\label{sec:simu} In this section we present a short simulation study to illustrate the usefulness of risk-RFE for feature elimination in KMs. Although risk-RFE is more useful in non-linear settings, we start off with a linear setting first to show that it can perform as well as notable linear methods like SCAD and LASSO in such a setting. And then, in the next part of this section, we move on to simple non-linear settings and demonstrate that risk-RFE dominates all the compared methods in such scenarios.

\subsection{Selection of features and consistency}\label{sec:simu_consis} The main aim of this section is to evaluate our consistency results through numerical examples, and a method for selection of the subset of features. We consider two different data-generating mechanisms, one in classification, and the other in regression. We look at four different scenarios (a) total number of covariates is $15$ of which $4$ are relevant, (b) $30$ covariates with $7$ relevant, (c) $50$ covariates with $3$ relevant, (d) $200$ covariates with $10$ relevant.
\begin{figure}[h!]
\begin{center}
\includegraphics[width=0.4\textwidth,height=0.25\textheight]{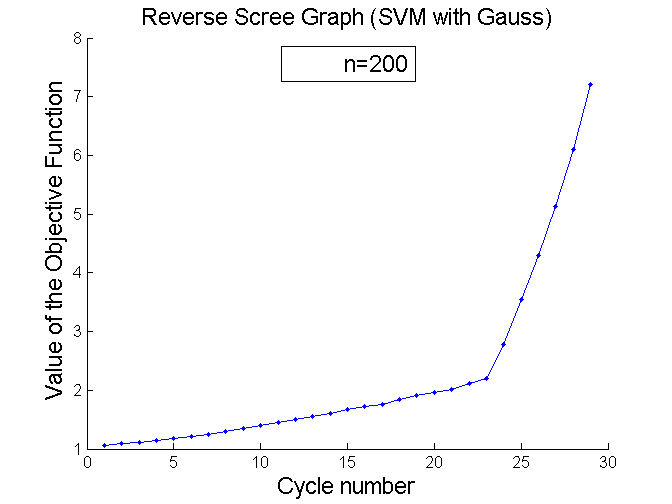}
\includegraphics[width=0.4\textwidth,height=0.25\textheight]{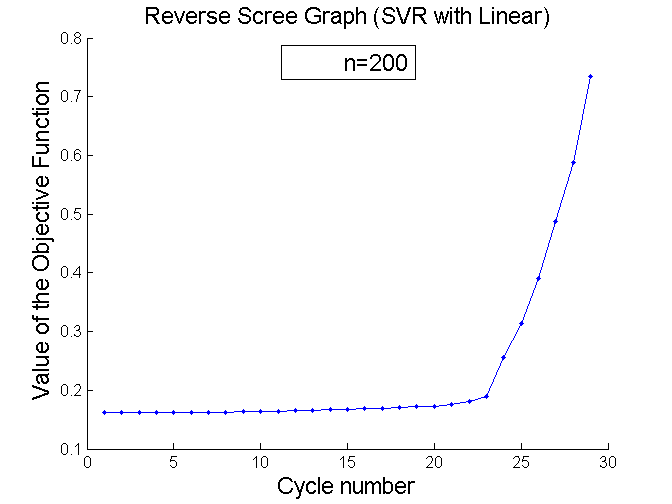}
\caption{Reverse Scree Graph for one run of the simulations for (a) SVM with Gaussian Kernel (b) SVR with Linear Kernel with $d=30$, $d_{0}=7$}
\label{fig:scree}
\end{center}
\end{figure}
For classification, we consider the hinge loss $L_{HL}$ as the surrogate replacement of the $0-1$ loss, and the KM function is computed using the Gaussian RBF kernel $k_{\gamma}(x_{1},x_{2})=\exp\{-\frac{1}{\gamma^{2}}\|x_{1}-x_{2}\|^{2}_{2}\}$. Covariates $X$ are generated uniformly on $\left[-1,\:1\right]$, and the output $Y$ is associated through a linear relationship with the covariate space $X$ given as $Y=\text{sign}(\omega^{\prime}X)$, where $\omega$ is the coefficient vector with only first few elements non-zero (corresponding to the relevant features chosen at random from a list of coefficients) and the rest are zero. The effect size of the relevant features are fixed to be either 0.5 or 1 except for the classification example in the high dimensional setting of $d=200$, $d_0=10$, where we enhance them by 50\% to get more meaningful comparisons. We initialize the original SVM function using a $5$-fold cross validation on the kernel width $\gamma$ and the regularization parameter $\lambda$ and they are chosen from the set of values $\left(\frac{2}{n\lambda},\gamma\right)=(0.01\times 10^{i},\:j),\: i=\{0,1,2,3,4\},\:j=\{1,2,3,4\}$, where $n$ is the sample size for the given setting.
%
\begin{figure}[t!]
\begin{center}
\includegraphics[width=0.4\textwidth,height=0.25\textheight]{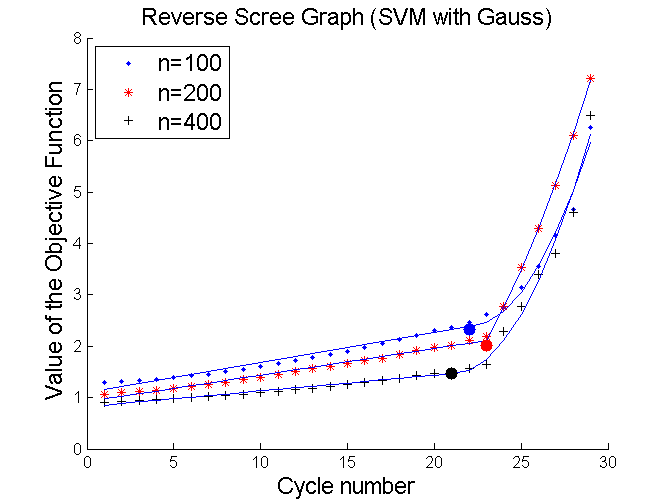}
\includegraphics[width=0.4\textwidth,height=0.25\textheight]{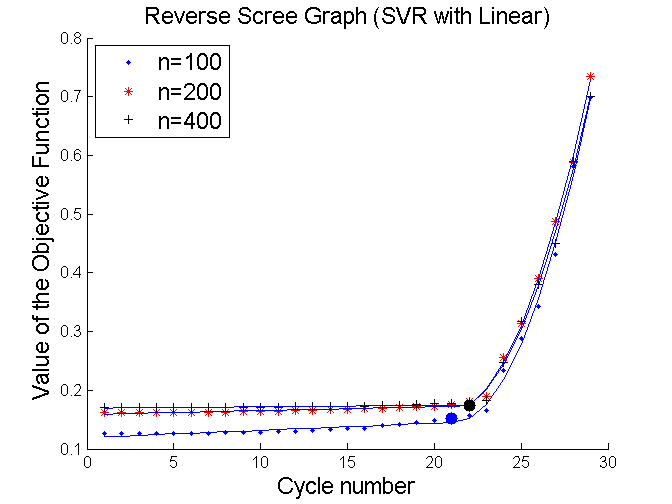}
\caption{Linear-Quadratic mixture change point analysis for (a) SVM with Gaussian Kernel for comparable cross validation values of $\lambda$ and kernel width $\gamma$ and (b) SVR with Linear Kernel for comparable cross validation values of $\lambda$, with $d=30$, $d_{0}=7$ for varying sample sizes. The bold dots represent the estimated change points.}
\label{fig:cp_30_7}
\end{center}
\end{figure}
\begin{figure}[t!]
\begin{center}
\includegraphics[width=0.4\textwidth,height=0.25\textheight]{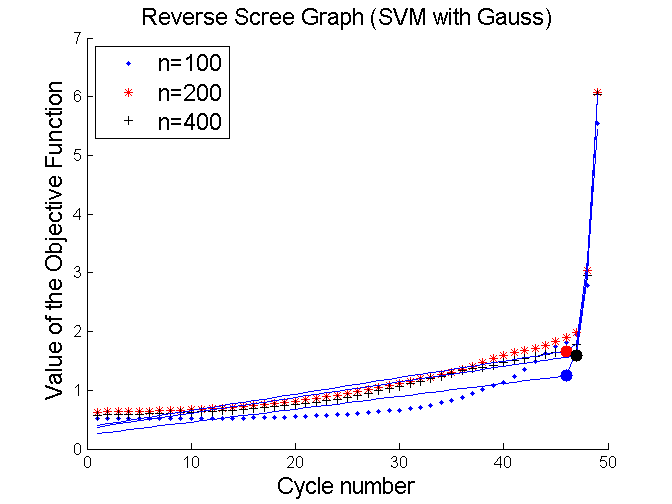}
\includegraphics[width=0.4\textwidth,height=0.25\textheight]{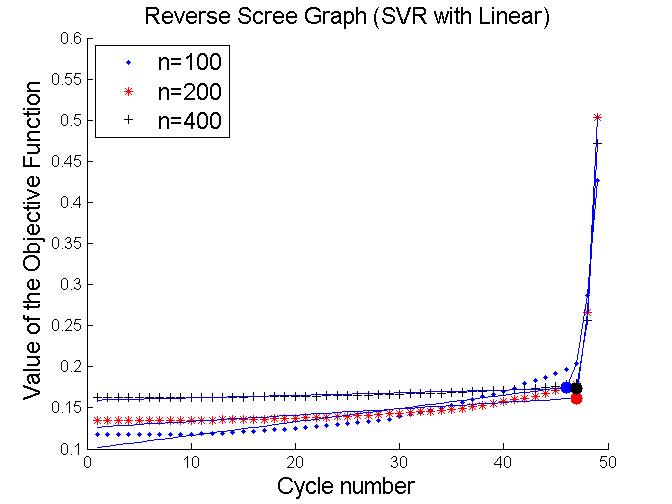}
\caption{Linear-Quadratic mixture change point analysis for (a) SVM with Gaussian Kernel for comparable cross validation values of $\lambda$ and kernel width $\gamma$ and (b) SVR with Linear Kernel for comparable cross validation values of $\lambda$, with $d=50$, $d_{0}=3$ for varying sample sizes. The bold dots represent the estimated change points.}
\label{fig:cp_50_3}
\end{center}
\end{figure}
We use a linear kernel $k(x_{1},x_{2})=\left\langle x_{1},x_{2} \right\rangle$ with the $\epsilon$-insensitive loss $L_{\epsilon}(x,y,f(x))=\max\{0,|y-f(x)|-\epsilon\}$  to treat the regression setting. The value of $\epsilon$ was fixed at $0.1$. Covariates are generated as before and $Y$ is still associated linearly with $X$, generated as $Y=\omega^{\prime}X+\frac{1}{3}N_{\dim(X)}(0,1)$. As before we initialize with a $5$-fold cross validation on $\lambda$. We repeat the procedures for different sample sizes $n=\{100,\:200,\:400\}$. We also repeat each simulation $100$ times. The entire methodology was implemented in MATLAB. For implementation of the SVM functions, we used the SPIDER library for MATLAB\footnote{The Spider library for Matlab can be downloaded from \url{http://www.kyb.tuebingen.mpg.de/bs/people/spider/}}, which already has a feature elimination algorithm based on Guyon's RFE, and we modified it to suit our criterion for deletion. The codes for the algorithm and the simulations are given in the \ref{sec:suppA}.

\begin{table}[h!]
\scriptsize
\setlength{\tabcolsep}{1pt}
\centering
\begin{tabular}{|l |cccc|cccc|cccc|}
\hline 
 \rule{0pt}{3ex}\textbf{Classification}   & \multicolumn{4}{c|}{$n=100$}&\multicolumn{4}{c|}{$n=200$}&\multicolumn{4}{c|}{$n=400$}\\[2pt]
\hline
\multirow{2}{*}{$d=15$, $d_0=4$} &\% no &\% 1&\% $>$1 & Mean &\% no &\% one &\% $>$1& Mean &\% no &\% 1 &\% $ >$1&Mean\\
 & error & error & error & Test Err &  error & error & error & Test Err & error & error & error & Test Err\\[2pt]
\hline
\rule{0pt}{3ex}SVM-G RRFE (NR) & 97 & 3 & 0 & 0.05& 100 & 0 & 0 & 0.03& 100 & 0 & 0 & 0.02\\
SVM-L RRFE (NR) & 99 & 1 & 0 & 0.04 & 100 & 0 & 0 & 0.02& 100 & 0 & 0 & 0.01\\
SVM-G GRFE & 17 & 38 & 45 & 0.20& 19 & 32 & 49 & 0.21& 32 & 21 & 47 & 0.18\\
SVM-L GRFE & 36 & 41 & 23 & 0.16& 61 & 27 & 12 & 0.11& 77 & 18 & 5 & 0.07\\
SCAD SVM & 98 & 2 & 0 & 0.04& 100 & 0 & 0 & 0.02& 100 & 0 & 0 & 0.01\\
Log Reg Lasso & 94 & 6 &0 & 0.03& 100 & 0 & 0 & 0.01& 100 & 0 & 0 & 0.01\\
\hline
\rule{0pt}{3ex}  \multirow{2}{*}{$d=30$, $d_0=7$} &\% no &\% 1&\% $>$1 & Mean &\% no &\% one &\% $>$1& Mean &\% no &\% 1 &\% $ >$1&Mean\\
 & error & error & error & Test Err &  error & error & error & Test Err & error & error & error & Test Err\\
\hline
\rule{0pt}{3ex}SVM-G RRFE (NR) & 62 & 34 & 4 & 0.09& 100 & 0 & 0 & 0.04& 100 & 0 & 0 & 0.03\\
SVM-L RRFE (NR) & 72 & 24 & 4 & 0.08 & 100 & 0 & 0 & 0.03& 100 & 0 & 0 & 0.02\\
SVM-G GRFE & 1 & 7 & 92 & 0.24 & 10 & 9 & 81 & 0.26 & 2 & 6 & 92 & 0.26\\
SVM-L GRFE & 10 & 24 & 75 & 0.23& 23 & 36 & 41 & 0.16& 55 & 32 & 13 & 0.10\\
SCAD SVM & 83 & 14 & 3 & 0.08& 100 & 0 & 0 & 0.03& 100 & 0 & 0 & 0.02\\
Log Reg Lasso & 46 & 49 & 5 & 0.09& 96 & 4 & 0 & 0.02& 100 & 0 & 0 & 0.01\\
\hline
\rule{0pt}{3ex}  \multirow{2}{*}{$d=50$, $d_0=3$} &\% no &\% 1&\% $>$1 & Mean &\% no &\% one &\% $>$1& Mean &\% no &\% 1 &\% $ >$1&Mean\\
 & error & error & error & Test Err &  error & error & error & Test Err & error & error & error & Test Err\\
\hline
\rule{0pt}{3ex}SVM-G RRFE (NR) & 95 & 5 & 0 & 0.05& 100 & 0 & 0 & 0.03& 100 & 0 & 0 & 0.02\\
SVM-L RRFE (NR) & 94 & 6 & 0 & 0.06 & 100 & 0 & 0 & 0.03& 100 & 0 & 0 & 0.01\\
SVM-G GRFE & 29 & 54 & 17 & 0.14 & 41 & 55 & 4 & 0.10 & 37 & 58 & 5 & 0.10\\
SVM-L GRFE & 41 & 30 & 29 & 0.20& 59 & 27 & 14 & 0.13& 89 & 9 & 2 & 0.04\\
SCAD SVM & 98 & 2 & 0 & 0.06& 100 & 0 & 0 & 0.03& 100 & 0 & 0 & 0.01\\
Log Reg Lasso & 97 & 3 &0 & 0.02 & 100 & 0 & 0 & 0.01& 100 & 0 & 0 & 0.01\\
\hline
\rule{0pt}{3ex} \multirow{2}{*}{$d=200$, $d_0=10$} &\% no &\% 1&\% $>$1 & Mean &\% no &\% one &\% $>$1& Mean &\% no &\% 1 &\% $ >$1&Mean\\
 & error & error & error & Test Err &  error & error & error & Test Err & error & error & error & Test Err\\
\hline
\rule{0pt}{3ex}SVM-G RRFE (NR) & 5 & 12 & 83 & 0.27& 72 & 24 & 4 & 0.07& 100 & 0 & 0 & 0.04\\
SVM-G RRFE (CP) & 10 & 25 & 65 & 0.21& 76 & 21 & 3 & 0.09& 100 & 0 & 0 & 0.02\\
SVM-L RRFE (NR) & 0 & 6 & 94 & 0.26 & 34 & 33 & 33 & 0.13& 97 & 3 & 0 & 0.05\\
SVM-L RRFE (CP) & 11 & 29 & 60 & 0.25	 & 56 & 37 & 7 & 0.10& 100 & 0 & 0 & 0.03\\
SVM-G GRFE & 0 & 0 & 100 & 0.42 & 0 & 0 & 100 & 0.40 & 2 & 3 & 95 & 0.36\\
SVM-L GRFE & 0 & 6 & 94 & 0.26& 30 & 39 & 31 & 0.15& 96 & 4 & 0 & 0.09\\
SCAD SVM & 3 & 18 & 79 & 0.24 & 69 & 26 & 5 & 0.11 & 100 & 0 & 0 & 0.05\\
Log Reg Lasso & 0 & 2 & 98 & 0.21 & 42 & 45 & 13 & 0.08& 99 & 1 & 0 & 0.02\\
\hline
\end{tabular}
\begin{center}
\caption{Comparison between (a) SVM-G RRFE - risk-RFE with SVM Gauss, (b)~SVM-L RRFE - risk-RFE with SVM Linear,
(c) SVM-G GRFE - Guyon RFE with SVM Gauss (d) SCAD SVM - Linear SVM with SCAD (e) Log Reg Lasso - Logistic Regression with LASSO in a classification setting (`NR' denotes using naive ranks, `CP' denotes using change point method) under linearity   %
    \label{tab:acc_rfe}}
\end{center}
\end{table}
One crucial question we face in feature elimination is when to stop. Note that Guyon's RFE has no inherent rule in this regard - it can only output the entire set of ranked features. One beautiful aspect of the risk-RFE algorithm is precisely that it can be used not only to rank features, but also for automatic selection of the optimal subset. This is achieved by noting this - in our theory, we proposed the existence of a gap $\epsilon_{0}$ and showed that asymptotically the empirical regularized risk of a model with at least one important feature missing exceeds that of a correct model by at least this amount. Practically it is very difficult to characterize this gap directly from the theory for a given setting, but its existence can be observed from plotting the objective function values at each iteration. Hence we can use the `Scree graph' of the objective function to build an auto-selection rule (for more on Scree graphs refer to chapter $6$ of \citet{Jolliffe2002}). 
\begin{figure}[h!]
\begin{center}
\includegraphics[width=0.5\textwidth,height=0.25\textheight]{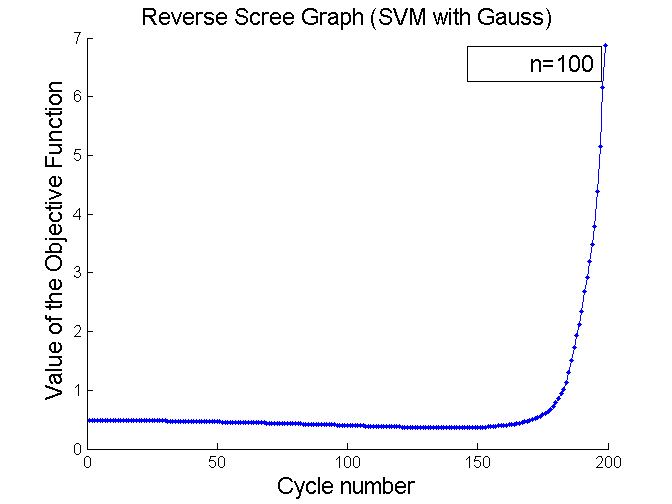}
\caption{Reverse Scree Graph for one run of the simulations for SVM with Gaussian Kernel with $d = 200$, $d_0 = 10$}
\label{fig:sg_200_10}
\end{center}
\end{figure}

Looking at Figure \ref{fig:scree}, it seems that a change-point model can be used for curve-fitting here, since we expect a change in the slope of the objective function as soon as we start eliminating significant features from the model (because of the aforementioned gap). Thus, one plausible way is to fit a change point regression model on the empirical objective function, and to infer the estimated change point as the ad-hoc stopping rule. For the belief that the change in the objective function is negligible to the left of the change point, so we fit a linear trend there. To the right however, these changes might show non-linear trends, and hence, we try both linear and quadratic trends to model them in our simulations. At least for the examples we tried, the quadratic trend seemed to work better. Figures \ref{fig:cp_30_7} and \ref{fig:cp_50_3} show our analysis where we use the mixture of linear-quadratic fits.

\begin{table}[h!]
\scriptsize
\setlength{\tabcolsep}{1pt}
\centering
\begin{tabular}{|l |cccc|cccc|cccc|}
\hline 
 \rule{0pt}{3ex} \textbf{Regression} & \multicolumn{4}{c|}{$n=100$}&\multicolumn{4}{c|}{$n=200$}&\multicolumn{4}{c|}{$n=400$}\\[2pt]
\hline
\multirow{2}{*}{$d=15$, $d_0=4$} &\% no &\% 1&\% $>$1 & Mean &\% no &\% one &\% $>$1& Mean &\% no &\% 1 &\% $ >$1&Mean\\
 & error & error & error & Test Err &  error & error & error & Test Err & error & error & error & Test Err\\[2pt]
\hline
\rule{0pt}{3ex}SVR-L RRFE & 100 & 0 & 0 &0.12 & 100 & 0 & 0 & 0.12& 100 & 0 & 0 & 0.11\\
SVR-L GRFE & 100 & 0 & 0 &0.12 & 100 & 0 & 0 & 0.12& 100 & 0 & 0 & 0.11\\
Lin Reg Lasso & 100 & 0 & 0 &0.12 & 100 & 0 & 0 & 0.12& 100 & 0 & 0 & 0.11\\
\hline
\rule{0pt}{3ex}  \multirow{2}{*}{$d=30$, $d_0=7$} &\% no &\% 1&\% $>$1 & Mean &\% no &\% one &\% $>$1& Mean &\% no &\% 1 &\% $ >$1&Mean\\
 & error & error & error & Test Err &  error & error & error & Test Err & error & error & error & Test Err\\
\hline
\rule{0pt}{3ex}SVR-L RRFE & 100 & 0 & 0 &0.13 & 100 & 0 & 0 & 0.12& 100 & 0 & 0 & 0.11\\
SVR-L GRFE & 100 & 0 & 0 &0.13 & 100 & 0 & 0 & 0.12& 100 & 0 & 0 & 0.11\\
Lin Reg Lasso & 99 & 1 & 0 &0.12 & 100 & 0 & 0 & 0.12& 100 & 0 & 0 & 0.11\\
\hline
\rule{0pt}{3ex}  \multirow{2}{*}{$d=50$, $d_0=3$} &\% no &\% 1&\% $>$1 & Mean &\% no &\% one &\% $>$1& Mean &\% no &\% 1 &\% $ >$1&Mean\\
 & error & error & error & Test Err &  error & error & error & Test Err & error & error & error & Test Err\\
\hline
\rule{0pt}{3ex}SVR-L RRFE & 100 & 0 & 0 &0.13 & 100 & 0 & 0 & 0.11& 100 & 0 & 0 & 0.11\\
SVR-L GRFE & 100 & 0 & 0 &0.13 & 100 & 0 & 0 & 0.11& 100 & 0 & 0 & 0.11\\
Lin Reg Lasso & 100 & 0 & 0 &0.12 & 100 & 0 & 0 & 0.11& 100 & 0 & 0 & 0.11\\
\hline
\rule{0pt}{3ex}  \multirow{2}{*}{$d=200$, $d_0=10$} &\% no &\% 1&\% $>$1 & Mean &\% no &\% one &\% $>$1& Mean &\% no &\% 1 &\% $ >$1&Mean\\
 & error & error & error & Test Err &  error & error & error & Test Err & error & error & error & Test Err\\
\hline
\rule{0pt}{3ex}SVR-L RRFE & 80 & 14 & 6 &0.22 & 100 & 0 & 0 & 0.13& 100 & 0 & 0 & 0.12\\
SVR-L GRFE & 81 & 14 & 5 &0.22 & 100 & 0 & 0 & 0.13& 100 & 0 & 0 & 0.12\\
Lin Reg Lasso & 49 & 31 & 20 &0.24 & 98 & 2 & 0 & 0.12& 100 & 0 & 0 & 0.11\\
\hline
\end{tabular}
\begin{center}
\caption{Comparison between (a)  SVM-L RRFE - risk-RFE with SVM Linear, (b)~SVM-G GRFE - Guyon RFE with SVM Gauss (c) Lin Reg Lasso - Linear Regression with LASSO in a regression setting (only `NR' was used)  under linearity %
    \label{tab:acc_rfe_reg}}
\end{center}
\end{table}
As a first confirmatory step, we decide to compare the performance of risk-RFE with that of Guyon's RFE to see if we indeed have enhanced performance in consistency of feature selection. We also compare it with other popular methods of feature selection in linear methods, namely SCAD SVM and logistic regression with LASSO in classification, and with linear regression with LASSO in regression. We present our results in Table \ref{tab:acc_rfe} (classification) and Table \ref{tab:acc_rfe_reg} (regression). We look at the percentage of times the algorithms (a) made no errors, (b) made only one error, or (c) made multiple errors, and also note the average test prediction error. Since the number of important features is known in each setting, instead of using the change point model, we take a more conservative approach, called the naive rank approach (NR) (this was also used for feature selection using Guyon's RFE) to calculate the errors for risk-RFE: a mistake is made if rank of any of the signals is lower than the total number of important features in the model. 

For the penalized methods, we use an ad-hoc cross validation technique to select the minimum amount of penalization that would result in retaining at least as many features as there are important features in the model. However, in the high dimensional case with $n/d=1/2$ ($d=200$, $d_0=10$, $n=100$), we use both the CP and NR formulations of risk-RFE to show that using the change point (CP) model can potentially do better at feature selection than using just the naive ranks, as the change point model can allow for better retention of signals at the cost of some additional noise in the model (see Figure \ref{fig:sg_200_10}).  

\begin{table}[h!]
\scriptsize
\setlength{\tabcolsep}{2pt}
\centering
\begin{tabular}{l ccccc}
\hline
\rule{0pt}{3ex}{\textbf{Method}}& \multicolumn{2}{c}{\textbf{Misclass Error}} & \multicolumn{3}{c}{\textbf{Consistency of selection}} \\[2pt]
&&& Prop. & Prop. & Prop.\\
& Mean & Std Err & all correct & 1 wrong & both wrong\\
\hline
SVM with risk-RFE (Gauss) & 0.06 & 0.02 & 1 & 0 & 0\\
SVM with Guyon-RFE (Gauss) & 0.07 & 0.05 & 0.94 & 0 & 0.06\\
Logistic Regression with Lasso & 0.25 & 0.04 &0.02 &0.3 & 0.68\\
SCAD SVM & 0.26 & 0.05 & 0.10 & 0 & 0.90\\
SVM without selection & 0.18 & 0.04 & na & na & na\\
\hline
\end{tabular}
\begin{center}
\caption{Non linear classification %
    \label{tab:class}}
\end{center}
\end{table}
Since the underlying relationship between $Y$ and the features $X$ is linear in each of the simulated examples in this section, we expect penalized methods to perform at least as well as risk-RFE in these settings. For the regression example, as can be seen in Table \ref{tab:acc_rfe_reg}, all methods are performing equally well in the lower dimensional settings, while LASSO is actually doing worse than both risk-RFE and Guyon's RFE when $n$ was small in the high dimensional setting ($d=200, d_0=10$). In the classification setting (see Table \ref{tab:acc_rfe}), Guyon's RFE is performing very poorly even when the sample size is high (using a linear kernel only improves the performance somewhat, but it is still quite poor), which justifies the need for the modification to get more consistent performances that we see from risk-RFE. Risk-RFE is dominating the performance of LASSO as well, more so in the smaller sample sizes, while in larger sample sizes both are performing equally well. Unsurprisingly, SCAD SVM is showing the best comparative performance here. Risk-RFE performs somewhat poorly in comparison to SCAD only in the setting $d=30, d_0=7, n=100$, while it does equally well as SCAD in all other settings. In fact, in the high dimensional setting ($d=200, d_0=10$) when $n=100$, using the change point method allowed risk-RFE to pick out signals with at most one mistake about 40\% of the times, compared to 21\% of the times that SCAD was able to do so. On further inspection of this setting, it was seen that when the CP method was used, risk-RFE allowed at most 5 mistakes overall (2-5 mistakes the remaining 60\% of the time). Risk-RFE with NR is actually doing similarly as SCAD SVM in this regard (both made 2-5 errors about 70\% of the time) in this scenario. A look at Table \ref{tab:acc_rfe} reveals that although logistic regression with LASSO is not always the best performing feature selection method, but it still obtains low  misclassification errors (in cases when its feature selection performance actually improves). This is probably due to the fact that the underlying relationships are all linear (in contrast to results in the next section where it obtains very poor misclassification errors), an effect the linear SVM couldn't replicate (as can be seen from errors obtained from the SCAD SVM and risk-RFE with Linear SVM). 

\begin{table}[h!]
\scriptsize
\setlength{\tabcolsep}{2pt}
\centering
\begin{tabular}{l ccccc}
\hline
\rule{0pt}{3ex}{\textbf{Method}}& \multicolumn{2}{c}{\textbf{Misclass Error}} & \multicolumn{3}{c}{\textbf{Consistency of selection}} \\[2pt]
&&& Prop. & Prop. & Prop.\\
& Mean & Std Err & all correct & 1 wrong & 2/3 wrong\\
\hline
SVM with risk-RFE (Gauss) & 0.28 & 0.14 & 0.78 & 0.12 & 0.10\\
SVM with Guyon-RFE (Gauss)& 0.28 & 0.14 & 0.8 & 0.12 & 0.08\\
Logistic Regression with Lasso & 0.49 & 0.03 &0.02 &0.02 & 0.96\\
SCAD SVM & 0.48 & 0.04 & 0 & 0 & 1\\
SVM without selection & 0.46 & 0.03 & na & na & na\\
\hline
\end{tabular}
\begin{center}
\caption{Non linear classification (presence of colinearity) %
    \label{tab:class2}}
\end{center}
\end{table}
%
\subsection{Risk-RFE in the non linear setup}\label{simu_lasso} As previously mentioned, risk-RFE becomes really powerful when the underlying functional form of the decision function is non-linear. In this section, we look at some simple non-linear settings in both classification and regression to ascertain the performance of risk-RFE in such scenarios. We consider two settings: (a) absence of colinearity or (b) presence of colinearity in covariates. In case (a), we generate ten features uniformly from $[-2,2]^{10}$, of which only two are important. In classification, $Y$ takes $1$ inside the smaller square $\left(-1\leq X_{1} \leq 1, -1\leq X_{2} \leq 1\right)$, and takes $-1$ inside the region formed between this square and the larger one given by $\left(-2\leq X_{1} \leq 2, -2\leq X_{2} \leq 2\right)$. In regression, $Y$ is given as $Y=\frac{a_{1}X_{1}X_{2}}{(1+a_{2}X_{1})^{2}}$, where $a_{1}$, $a_{2}$ are strictly positive constants. In case (b), we generate fifty features uniformly from $[-2^{\frac{1}{3}},2^{\frac{1}{3}}]^{50}$ of which only three are important, and these are either positively or negatively correlated amongst themselves. In classification, we define $Y$ to be $1$ inside the cube $\left(-1\leq X_{1} \leq 1, -1\leq X_{2} \leq 1,-1\leq X_{3} \leq 1\right)$, and $-1$ inside the region formed between this cube and the larger cube given as $\left(-2^{\frac{1}{3}}\leq X_{1} \leq 2^{\frac{1}{3}}, -2^{\frac{1}{3}}\leq X_{2} \leq 2^{\frac{1}{3}},-2^{\frac{1}{3}}\leq X_{3} \leq 2^{\frac{1}{3}}\right)$. In regression, $Y$ is now given as $Y=\frac{a_{1}X_{1}X_{2}}{(1+a_{2}X_{1})^{2}}+\frac{a_{3}X_{2}X_{3}}{(1+a_{4}X_{2})^{2}}$, where $a_{1}$, $a_{2}, a_3$ and $a_4$ are strictly positive constants.

\begin{table}[h!]
\scriptsize
\setlength{\tabcolsep}{2pt}
\centering
\begin{tabular}{l ccccc}
\hline
\rule{0pt}{3ex}{\textbf{Method}}& \multicolumn{2}{c}{\textbf{Mean Squared Error}} & \multicolumn{3}{c}{\textbf{Consistency of selection}} \\[2pt]
&&& Prop. & Prop. & Prop.\\
& Mean & Std Err & all correct & 1 wrong & both wrong\\
\hline
SVM with risk-RFE (Gauss)& 0.38 & 0.20 & 1 & 0 & 0\\
SVM with Guyon-RFE (Gauss)& 0.38 & 0.20 & 1 & 0 & 0\\
Linear Regression with Lasso & 1.90 & 0.55 &0 &0.44 & 0.56\\
SVM without selection & 0.65 & 0.22 & na & na & na\\
\hline
\end{tabular}
\begin{center}
\caption{Non linear regression %
    \label{tab:reg}}
\end{center}
\end{table}
In classification, we compare risk-RFE with Guyon's RFE, SCAD SVM and logistic regression with LASSO, and in regression, we compare risk-RFE with Guyon's RFE and linear regression with LASSO. In all scenarios, we use the Gaussian RBF kernel to train the SVM for running the RFE algorithms, and use the naive ranks for obtaining the correct set for all of them. The results for classification are given in tables \ref{tab:class} and \ref{tab:class2}, while those for regression are given in \ref{tab:reg} and \ref{tab:reg2}. As suspected, risk-RFE is doing substantially better than SCAD and LASSO in all scenarios (not surprisingly both SCAD and LASSO perform very poorly in all scenarios they were used in), and was able to pick the correct features consistently in most cases. Guyon's RFE is doing relatively well as well, owing to the fact that it is applicable in the non-linear setting as well. Guyon's RFE is doing slightly better than risk-RFE in classification under colinearity, while doing very poorly under colinearity in the regression case. On the other hand, performance of risk-RFE did not drop too much across examples. 

\begin{table}[h!]
\scriptsize
\setlength{\tabcolsep}{2pt}
\centering
\begin{tabular}{l ccccc}
\hline
\rule{0pt}{3ex}{\textbf{Method}}& \multicolumn{2}{c}{\textbf{Mean Squared Error}} & \multicolumn{3}{c}{\textbf{Consistency of selection}} \\[2pt]
&&& Prop. & Prop. & Prop.\\
& Mean & Std Err & all correct & 1 wrong & 2/3 wrong\\
\hline
SVM with risk-RFE (Gauss)& 0.90 & 0.14 & 1 & 0 & 0\\
SVM with Guyon-RFE (Gauss)& 1.40 & 0.32 & 0.12 & 0.86 & 0.02\\
Linear Regression with Lasso & 2.24 & 0.14 &0 &0 & 1\\
SVM without selection & 1.40 & 0.09 & na & na & na\\
\hline
\end{tabular}
\begin{center}
\caption{Non linear regression (presence of colinearity) %
    \label{tab:reg2}}
\end{center}
\end{table}
%
\section{Discussion}\label{sec:dis} We proposed an algorithm for feature elimination in kernel machines. We studied the theoretical properties of this method, and showed that it is consistent in finding the correct feature space, when the functional space has certain special properties (nestedness or denseness). We then discussed additional assumptions that become necessary for a recursive
algorithm to become useful in a general functional space, and then established consistency for our method under this most general setting. We provided four case studies presenting different scenarios where this method can be used. We provided a short simulation study to illustrate the method and discuss a practical method for choosing the correct subset of features. Finally, we discussed the case when the dimension of the input space grows with the sample size.

We established the existence of a gap in the rate of change of the objective function at the point where our feature elimination method starts removing the essential features of the learning problem. This motivated us to use a Scree plot of the values of the objective function at each iteration, and indeed our simulation results support our approach by visually exhibiting this gap in the plots. Moreover, the graphical interpretation of the scree plot motivated the use of change point regression to select the correct feature space. It would thus be interesting to conduct a more detailed and formal analysis of this gap in real life settings to facilitate more efficient and automated practical solutions.


From our discussion in Section \ref{sec:large_p}, when dimension $d$ grows with $n$, we saw that risk-RFE is most effective when used under certain restrictions on the design size $d$ relative to the sample size $n$. One immediate question that arises then is if it can be modified further to be used in `large $p$, small $n$' problems, and this remains a topic of interest to us. 
Many of the current state of the art ultra-high feature elimination techniques depend on the correlation structure between the input-output space, such as the sure independence screening (SIS) method of \citet{Fan2007}, or other SIS based methods such as DC-SIS \citep{Li2012}.
Typically in linear models, these methods are used to effectively screen models from ultra high dimensions to a lower dimensional setting, wherein more meaningful lower dimensional methods like SCAD and LASSO become applicable. Hence, we propose that when dealing with ultra high dimensions in KM, risk-RFE be used in conjunction with one of these methods. As we saw, if the underlying model is non-linear, using risk-RFE after initial feature screening would really enhance the performance of the KM function compared to other available techniques.

As far as we are aware, not much analysis has been done on the properties of variable selection algorithms under such general transformations of the input space in kernel machines. Hence, the results generated in this paper can act as a good starting point for similar analyses in other settings. 
It would also be interesting to analyze RFE in censored support vector regression (See \citet{goldberg2013}) and other machine learning problems (including reinforcement learning) or other penalized risk minimization problems.

\begin{appendix}

\section{Functional Spaces on Lower Dimensional Domains}\label{sec:featel} The aim of this section is to provide some results on the restricted spaces defined in Definition \ref{defn:fj} in Section \ref{sec:feat_algo}.

\subsection{Further discussions on the lower dimensional spaces $\mathcal{F}^{J}$}\label{sec:fj_hj} 
We provide a few results here that connect the restricted functional spaces with the original one. In view of Definition~\ref{defn:fj}, we can define $\mathcal{L}_{\infty}^{J}(\mathcal{X})=\{f\circ\pi^{J^{c}}\::\:f\in \mathcal{L}_{\infty}(\mathcal{X})\}$. Then Lemma~\ref{lem:linfty} below says that $\mathcal{L}_{\infty}^{J}(\mathcal{X}^{J})\equiv \mathcal{L}_{\infty}^{J}(\mathcal{X})\big|_{\mathcal{X}^{J}}$ is isomorphic to the space $\mathcal{L}_{\infty}(\mathcal{X}^{J})$. Lemma~\ref{lem:handhj} below, provides some results connecting the original RKHS with its lower dimensional versions. A related lemma, Lemma \ref{lem:fandfj}, is given in the \ref{sec:suppA} noting similar results for any general space. These results aim to show that many of the nice properties of a given functional space are carried forward to their re-adaptations under Definition \ref{defn:fj}. We prove Lemma \ref{lem:linfty} and \ref{lem:fandfj}, while the proof for Lemma \ref{lem:handhj} is omitted as it follows from Lemma \ref{lem:fandfj} trivially. The proofs can be found in the \ref{sec:suppA}. 

\begin{lem}\label{lem:linfty} $\mathcal{L}_{\infty}^{J}(\mathcal{X}^{J})=\mathcal{L}_{\infty}(\mathcal{X}^{J})$.
\end{lem}

\begin{lem}\label{lem:handhj} Let $H \subset \mathcal{L}_{\infty}(\mathcal{X})$ be a non-empty RKHS on $\mathcal{X}$. Then for any $J \subset \{1,2,\dots,d\}$,
\begin{enumerate}
	\item If $H$ is dense in $\mathcal{L}_{\infty}(\mathcal{X})$, then $H^{J}$ is dense in $\mathcal{L}_{\infty}(\mathcal{X}^{J})$.\label{itm:dense_svm}
	\item If the $\|\cdot\|_{\infty}$ closure $\overline{B_{H}}$ of the unit ball $B_{H}$ is compact, then so is $\overline{B_{H^{J}}}$.\label{itm:comp_svm}
	\item If $H$ is separable, then so is $H^{J}$.\label{itm:sep_svm}
	\item $e_{i}(id:H^{J}\mapsto L_{\infty}(\mathcal{X}))$ $\leq$ $e_{i}(id:H\mapsto L_{\infty}(\mathcal{X}))$, where $e_{i}(id:H\mapsto L_{\infty}(\mathcal{X}))$ is the $i^{th}$ entropy number of the unit ball $B_{H}$ of the RKHS $H$, with respect to the $\|\cdot\|_{\infty}$-norm.\label{itm:entropy_bound}
\end{enumerate}
\end{lem}	
NOTE: \ref{itm:dense_svm} holds true even if $\mathcal{L}_{\infty}(\mathcal{X})$ is replaced by any space that admits the nested structure.

%

\subsection{RKHS in lower dimensions}\label{sec:hj1_hj2} Note that in kernel machines, the minimization is computed over an RKHS, and hence, while defining these lower dimensional spaces we need to ensure that these spaces are RKHSs as well. To that effect, we begin this section by providing an alternate way to define the lower dimensional versions of a given RKHS that preserves the reproducing property.

\begin{defn}\label{defn:hj}
For a given RKHS $H$ indexed by a kernel $k$ and a set of indices $J \subseteq \{1,2,..,d\}$, define $H^{J}\equiv H_{k\circ\pi^{J^{c}}}(\mathcal{X})$, where $k\circ\pi^{J^{c}}(x,y):=k(\pi^{J^{c}}(x),\pi^{J^{c}}(y))$.
\end{defn}

Note immediately that Definition \ref{defn:hj} allows us to create lower dimensional versions of $H$ in a way which ensures that these spaces are RKHS as well. This inevitably raises questions about the validity of Definition \ref{defn:fj} from section \ref{sec:feat_algo}. However both Definitions \ref{defn:fj} and \ref{defn:hj} yield the same RKHS space $H^{J}$.

To see this, let $\mathcal{X}_{0}$ be a subset of $\mathcal{X}$ and $k^{(0)}(x,y)$ be the restriction of a kernel $k$ on $\mathcal{X}_{0}$. Then $k^{(0)}(x,y)$ is a valid kernel on $\mathcal{X}_{0}$ by Proposition 5.13 of \citet{Paulsen2009}.
Let $H_{k}(\mathcal{X})$ be the RKHS with respect to $k(x,y)$, and $H_{k^{(0)}}(\mathcal{X})$ be the one with respect to $k^{(0)}(x,y)$. Then by Theorem 5.14 of  \citet{Paulsen2009}, if we define $\varphi$ to be the inclusion id map from $\mathcal{X}_{0}$ to $\mathcal{X}$, we have
$H_{k^{(0)}}(\mathcal{X}_{0})=\{f|_{\mathcal{X}_{0}}:f \in H_{k}(\mathcal{X})\}$ and $\|g\|_{H_{k^{(0)}}}=\min\{\|f\|_{H_{k}}\::\:f|_{\mathcal{X}_{0}}=g\}$
for $g \in H_{k^{(0)}}(\mathcal{X}_{0})$. Taking $\mathcal{X}_{0} \equiv \mathcal{X}^{J}$ and $k^{(0)}(x,y) \equiv k(\pi^{J^{c}}(x),\pi^{J^{c}}(y))$, we immediately obtain our assertion. 

\section{Further discussion on Case Study 2}\label{sec:cs2_proof}

\subsection{Proof of Lemma \ref{lem:rbf_delta} (from section \ref{sec:cs2})} In order to prove Lemma \ref{lem:rbf_delta} for the Guassian RBF kernel $k_{\gamma}(x,y)=e^{-\gamma^{2}\|x-y\|^{2}_{2}}$, we need to verify the regularity conditions given before Theorem \ref{thm:main2} from section \ref{sec:consis} in this setup. First note that $L_{HL}$ is Lipschitz continuous and bounded for all $3$-tuples of the form $(x,y,0)$ (see Example $2.27$ in \citetalias{SVR}). Separability of $H_{\gamma}$ holds since an RKHS over a separable metric space having a continuous kernel is separable (Lemma $4.33$ of \citetalias{SVR}). It is also easy to see that $|k_{\gamma}(x,y)|\leq 1$ is true for all $x, y \in\mathcal{X}$ and all $\gamma > 0$ and hence $\|k_{\gamma}\|_{\infty}\leq 1$.

From the proof of Proposition~\ref{prop:svm} in section \ref{sec:prop_svm_proof} we can see that the assumption on the bound on the average entropy of the RKHS space given before Theorem \ref{thm:main2}, can be replaced by the following:
\begin{itemize}
\item\label{as:entropy2} We assume that for fixed $n\geq1$, $\exists$ constants $a\geq1$ and $p\in (0,1)$ such that for any $J \subseteq \{1,2,\dots,d\}$,
$	\mathbb{E}_{D_{\mathcal{X}}\sim P^{n}_{\mathcal{X}}}e_{i}\left(id:H^{J}\mapsto L_{2}(D_{\mathcal{X}})\right) \leq ai^{-\frac{1}{2p}},\quad i \geq 1$.
\end{itemize}
It is easily seen from the steps in \eqref{eq:linf_ent_bound} in Appendix \ref{sec:prop_svm_proof} that results will hold if we replace the earlier assumption with the latter. Then we see that Theorem $7.34$ with Corollary $7.31$ of \citetalias{SVR} along with the fact that $d/(d+\tau)$ is an increasing function in $d$, yields a bound as given here with $a:=c_{\epsilon,p}\gamma^{\frac{(1-p)(1+\epsilon)d}{2p}}$ for all $\gamma\geq 1$, for all $\epsilon>0$, $d/(d+\tau)<p<1$ and a constant $c_{\epsilon,p}$ depending only on $p$ and a given $\epsilon$, and where $\tau\in \left( \right. 0,\infty\left.\right]$ is the tail exponent of the distribution $P_\mathcal{X}$ \citepalias[see chapter 7 of ][for definition of the tail exponent of a distribution]{SVR}.

Now we need to bound $A_2^{J_{\ast}}(\lambda)$. Note that this can be obtained from Theorem $2.7$ in \citet{Steinwart07},
where we see that the approximation error for a SVM using Gaussian RBF kernel of width $1/\gamma$ and with number of signals $d_0$ can be bounded by
\begin{align}
A_{2}(\lambda,d_0,\gamma)\leq c_{d_0} \left(\gamma^{d_0}\lambda + C (2d_0)^{\alpha d_0/2}\gamma^{-\alpha d_0}\right),
\end{align}
where $P$ is a distribution on $\mathcal{X}^{J_{\ast}} \times \{-1, 1\} \subset \mathbb{R}^{d_0}\times \{-1, 1\}$ where $\alpha \in (0,\infty)$ is geometric noise exponent and $C$ is the constant appearing in Definition $2.2$ of \citet{Steinwart07}, and $c_{d_0}$ is also a constant depending only on $d_0$. So for a given pair $(\lambda,\: d_0)$ if we choose $\gamma(\lambda,d_0)=\lambda^{-\frac{1}{(\alpha+1)d_0}}$ then it can be seen that $A_{2}(\lambda,d_0,\gamma(\lambda,d_0))\leq K_{d_0} \lambda^{\frac{\alpha}{\alpha+1}}$.

So for a sequence of SVM objective functions $\lambda_{n}\|f\|^{2}_{H_{\gamma(\lambda_{n})}}+\frac{1}{n}\sum_{i=1}^{n}\max\{0, 1-y_{i}f(x_{i})\}$ defined for a sequence $\lambda_{n}^{-1}=o(n)$ with $\lambda_{n}\rightarrow 0$ the assumptions for the theoretical results on consistency of RFE are met, and thus Lemma \ref{lem:rbf_delta} is proved.

\section{Proofs}\label{sec:appendA}

\subsection{Proof of Lemma~\ref{lem:rd-rp} (from section \ref{sec:theory})}\label{sec:rd-rp_proof}
\begin{proof}
Note that if we define $g_{f}:=L\circ f-E_{P}(L\circ f)$, then $\mathcal{G}=\{g_{f}:f\in \mathcal{F}\}$ is a separable Carath$\acute{e}$odory set (for a discussion on Carath$\acute{e}$odory families of maps, refer to Definition $7.4$ in \citetalias{SVR}). To see this, first note that $\|g_{f}\|_{\infty}\leq \displaystyle\sup_{(x,y)\in \mathcal{X}\times\mathcal{Y}}\left|L\circ f-E_{P}(L\circ f)\right|\leq 2B$ for $B$ defined in the statement of the Lemma. Also by assumption, $\|\cdot\|_{\mathcal{F}}$ dominates the pointwise convergence of functions (so $f_{n}\rightarrow f$ in $\|\cdot\|_{\mathcal{F}}$ $\Rightarrow$ $f_{n}\rightarrow f$ pointwise). Then the fact that $L$ is locally-Lipschitz continuous coupled with Lebesgue's Dominated Convergence Theorem (since $\|L\circ f\|_{\infty}\leq B$) gives us the above assertion.

Now note that $E_{P}(g_{f})=0$ and $E_{P}g_{f}^{2}\leq (2B)^{2}=4B^{2}$ for $B$ as before, so we can apply Talagrand's Inequality as given in Theorem $7.5$ of \citetalias{SVR} on $G$ defined as $G:\mathcal{Z}^{n}\equiv(\mathcal{X}\times\mathcal{Y})^{n}\mapsto \mathbb{R}$ such that
{\small\begin{align}
G(z_{1},\dots,z_{n}):=\displaystyle\sup_{g_{f}\in\mathcal{G}}\left|\frac{1}{n}\displaystyle\sum_{j=1}^{n}g_{f}(z_{j})\right|=\displaystyle\sup_{f\in\mathcal{F}}\left|\mathcal{R}_{L,D}(f)-\mathcal{R}_{L,P}(f)\right|,
\end{align}}%
and hence, for $\gamma =1$ and for all $\tau> 0$, we have
{\small\begin{align}
P^{n}\left(\left\{z\in \mathcal{Z}^{n}:G(z)\geq 2E_{P^{n}}(G)+2B\sqrt{\frac{2\tau}{n}}+\frac{10B\tau}{3n}\right\}\right)\leq e^{-\tau}.
\end{align}}%
So now we need to bound the term $E_{P^{n}}(G):=E_{P^{n}}\left\{\displaystyle\sup_{f\in\mathcal{F}}\left|\mathcal{R}_{L,D}(f)-\mathcal{R}_{L,P}(f)\right|\right\}$.

Defining the new Carath$\acute{e}$odory set $\mathcal{H}$ as $\mathcal{H}=\{h_{f}:=L\circ f\::\:f\in \mathcal{F}\}$, for a probability distribution $P$ on $\mathcal{Z}\equiv(\mathcal{X}\times\mathcal{Y})$, we can use the idea of symmetrization given in Proposition $7.10$ in \citetalias{SVR} to bound $E_{P^{n}}\left\{\displaystyle\sup_{f\in\mathcal{F}}\left|\mathcal{R}_{L,D}(f)-\mathcal{R}_{L,P}(f)\right|\right\}$. We have for all $n\geq 1$,
{\small\begin{align*}
E_{D\sim P^{n}}\left\{\displaystyle\sup_{f\in\mathcal{F}}\left|\mathcal{R}_{L,D}(f)-\mathcal{R}_{L,P}(f)\right|\right\} &	= E_{D\sim P^{n}}\displaystyle\sup_{h_{f}\in \mathcal{H}}\left|E_{P}h_{f}-E_{D}h_{f}\right| \\ & \leq 2E_{D\sim P^{n}}\text{Rad}_{D}(\mathcal{H},n),
\end{align*}}%
where $\text{Rad}_{D}(\mathcal{H},n)$ is the $n$-th empirical Rademacher average of $\mathcal{H}$ for $D:=\{z_{1},\dots,z_{n}\}\in \mathcal{Z}^{n}$ with respect to the Rademacher sequence $\{\varepsilon_{1},\dots,\varepsilon_{n}\}$ and the distribution $\nu$, which is given by $\text{Rad}_{D}(\mathcal{H},n)=E_{\nu}\displaystyle\sup_{h\in \mathcal{H}}\left|\frac{1}{n}\displaystyle\sum_{i=1}^{n}\varepsilon_{i}h(z_{i})\right|$. So we see now that it suffices to bound $E_{D\sim P^{n}}\text{Rad}_{D}(\mathcal{H},n)$.

For that we use theorem $7.16$ of \citetalias{SVR}, but before that note that the entropy bound means we have for fixed $n\geq1$, that $\exists$ constants $a\geq1$ and $p\in (0,1)$ such that
{\small\begin{align}
\mathbb{E}_{D_{\mathcal{X}}\sim P^{n}_{\mathcal{X}}}e_{i}\left(\mathcal{F}, L_{\infty}(D_{\mathcal{X}})\right) \leq ai^{-\frac{1}{2p}},\quad\quad i \geq 1.
\end{align}}%
First observe that $\mathcal{H}\subset\mathcal{L}_{2}(P)$. Now since Lipschitz continuity of $L$ gives us that $|L(x,y,f_{1}(x))-L(x,y,f_{2}(x))|^{2}\leq c_{L}(C)^{2}|f_{1}(x)-f_{2}(x)|^{2}$, it is easy to see that\\ $e_{i}(\mathcal{H},\|\cdot\|_{L_{2}(P)})\leq c_{L}(C)e_{i}(\mathcal{F},\|\cdot\|_{L_{2}(P_{\mathcal{X}})})$. Hence we have
{\small\begin{align}\label{eq:linf_ent_bound}
E_{D\sim P^{n}}\left(e_{i}(\mathcal{H},\|\cdot\|_{L_{2}(D)})\right) &\leq  c_{L}(C)E_{D_{\mathcal{X}}\sim P^{n}_{\mathcal{X}}}\left(e_{i}(\mathcal{F},\|\cdot\|_{L_{2}(D_{\mathcal{X}})})\right)\\ &\leq  c_{L}(C)E_{D_{\mathcal{X}}\sim P^{n}_{\mathcal{X}}}\left(e_{i}(\mathcal{F},\|\cdot\|_{L_{\infty}(D_{\mathcal{X}})})\right)\nonumber\\&\leq c_{L}(C)ai^{-\frac{1}{2p}}.\nonumber
\end{align}}%
Now noting that $\|h_{f}\|_{\infty}\leq B$ and $E_{P}h_{f}^{2}\leq B^{2}$ for $B$ defined as before, the conditions of Theorem $7.16$ of \citetalias{SVR} are satisfied with $\widetilde{a}=c_{L}(C)a$ and hence we have,
{\small\begin{align}
E_{D\sim P^{n}}\text{Rad}_{D}(\mathcal{H},n)\leq \max\left\{C_{1}(p)\widetilde{a}^{p}B^{1-p}n^{-\frac{1}{2}},C_{2}(p)\widetilde{a}^{\frac{2p}{1+p}}B^{\frac{1-p}{1+p}}n^{-\frac{1}{1+p}}\right\}
\end{align}}%
for constants $C_{1}(p)$, $C_{2}(p)$ depending only on $p$. Hence we finally have that with probability $\geq 1 - e^{-\tau}$,
{\small\begin{eqnarray*}
\lefteqn{\displaystyle\sup_{f \in \mathcal{F}} \left|\mathcal{R}_{L,P}(f)-\mathcal{R}_{L,D}(f)\right| \leq 2B\sqrt{\frac{2\tau}{n}}+\frac{10B\tau}{3n}}&&\\ &+& 4\max\left\{C_{1}(p)c_{L}(C)^{p}a^{p}B^{1-p}n^{-\frac{1}{2}},C_{2}(p)c_{L}(C)^{\frac{2p}{1+p}}a^{\frac{2p}{1+p}}B^{\frac{1-p}{1+p}}n^{-\frac{1}{1+p}}\right\}.
\end{eqnarray*}}%

\end{proof}

\subsection{Proof of Proposition~\ref{prop:svm} (from section \ref{sec:theory})}\label{sec:prop_svm_proof}
\begin{proof}
First note that since $B\geq 1$ and $K\geq B^{p}/4$, we have $24KB^{1-p}\geq 6B > 2$. Recall the notational definitions \eqref{eq:shorts} from section \ref{sec:main2_proof} and then see that if $a^{2p}>\lambda^{p}n$, the inequality trivially follows from the fact that
{\small\begin{eqnarray*}
&&\left|\mathcal{R}\mathcal{R}^D_D\left(J_1,J_2\right)\right| \leq \left|\lambda\left\|f_{D,\lambda,H^{J_{2}}}\right\|^{2}_{H^{J_{2}}} + \mathcal{R}_{L,D}\left(f_{D,\lambda,H^{J_{2}}}\right)\right|\\ &&\quad + \left|\lambda\left\|f_{D,\lambda,H^{J_{1}}}\right\|^{2}_{H^{J_{1}}} + \mathcal{R}_{L,D}\left(f_{D,\lambda,H^{J_{1}}}\right)\right|\\
&&\leq 2\mathcal{R}_{L,D}(0)\leq 24KB^{1-p}\left(\frac{a^{2p}}{\lambda^{p}n}\right)^{\frac{1}{2}},
\end{eqnarray*}}%
since $\mathcal{R}_{L,D}(0)\leq 1$. Hence we assume from here on that $a^{2p}\leq \lambda^{p}n$.
Now observe that since $H$ is separable, from Lemma \ref{lem:handhj} in section \ref{sec:featel} we have that the $H^{J}$s are also separable. Hence from Lemma $6.23$ of \citetalias{SVR} we have that the KMs produced by these RKHSs are measurable.

Now note that $L(x,y,0)\leq 1$ $\Rightarrow$ for any distribution $Q$ on $\mathcal{X}\times\mathcal{Y}$, we have that $\mathcal{R}_{L,Q}(0) \leq 1$. Since, $\displaystyle\inf_{f \in H^{J}} \lambda\|f\|^{2}_{H^{J}}+\mathcal{R}_{L,Q}(f)\leq \mathcal{R}_{L,Q}(0)$, we have that $\left\|f_{Q,\lambda,H^{J}}\right\|_{H^{J}}\leq \sqrt{\frac{\mathcal{R}_{L,Q}(0)}{\lambda}}$. Now since by Lemma $4.23$ of \citetalias{SVR} $\|f\|_{\infty}\leq\|k\|_{\infty}\|f\|_{H^{J}}$ for all $f \in H^{J}$, we have that $\left\|f_{Q,\lambda,H^{J}}\right\|_{\infty}\leq\left\|f_{Q,\lambda,H^{J}}\right\|_{H^{J}}\leq\lambda^{-1/2}$. So, consequently, for every distribution $Q$ on $\mathcal{X}\times\mathcal{Y}$, we have
{\small\begin{align}\label{eq:emp_bound1}
\left|\mathcal{R}_{L,P}\left(f_{Q,\lambda,H^{J}}\right)-\mathcal{R}_{L,D}\left(f_{Q,\lambda,H^{J}}\right)\right| \leq \displaystyle\sup_{\|f\|_{H^{J}} \leq \lambda^{-1/2}} \left|\mathcal{R}_{L,P}(f)-\mathcal{R}_{L,D}(f)\right|.
\end{align}}%
Now since from \ref{as:a1} in section \ref{sec:disc_assump}, $\mathcal{R}^{\ast}_{L,P,H^{J_{1}}}=\mathcal{R}^{\ast}_{L,P,H^{J_{2}}}=\mathcal{R}^{\ast}_{L,P,H}$, we have 
{\small\begin{align*}
&\left|\mathcal{R}\mathcal{R}^D_D\left(J_1,J_2\right)\right| \leq \left|\mathcal{R}\mathcal{R}^D_P\left(J_2\right)\right| +\left|\mathcal{R}\mathcal{R}^D_P\left(J_1\right)\right| +\left|\mathcal{R}_{L,D}\left(f_{D,\lambda,H^{J_{2}}}\right)-\mathcal{R}_{L,P}\left(f_{D,\lambda,H^{J_{2}}}\right)\right|\\
&+ \left|\mathcal{R}_{L,D}\left(f_{D,\lambda,H^{J_{1}}}\right)-\mathcal{R}_{L,P}\left(f_{D,\lambda,H^{J_{1}}}\right)\right|.
\end{align*}}%
Noting that $\mathcal{R}\mathcal{R}^D_P\left(J\right) \geq 0$, we have from $(6.18)$ of \citetalias{SVR}
{\small\begin{align}\label{eq:emp_bound2}
&\left|\mathcal{R}\mathcal{R}^D_P\left(J\right)\right|\leq A_{2}^{J}(\lambda)+\mathcal{R}_{L,P}\left(f_{D,\lambda,H^{J}}\right)-\mathcal{R}_{L,D}\left(f_{D,\lambda,H^{J}}\right)\\
&\quad+\mathcal{R}_{L,D}\left(f_{P,\lambda,H^{J}}\right)-\mathcal{R}_{L,P}\left(f_{P,\lambda,H^{J}}\right) \leq A_{2}^{J}(\lambda)+2\displaystyle\sup_{\|f\|_{H^{J}} \leq \lambda^{-1/2}} \left|\mathcal{R}_{L,P}(f)-\mathcal{R}_{L,D}(f)\right|.\nonumber
\end{align}}%
From \eqref{eq:emp_bound1} and \eqref{eq:emp_bound2} and the fact that $J_{1},\:J_{2} \in\widetilde{\mathcal{J}}$ such that $J_{1} \subseteq J_{2} \subseteq J_{\ast}$, we have that
{\small\begin{align*}
&\left|\mathcal{R}\mathcal{R}^D_D\left(J_1,J_2\right) \right| \leq 3\displaystyle\sup_{\|f\|_{H^{J_{1}}} \leq \lambda^{-1/2}} \left|\mathcal{R}_{L,P}(f)-\mathcal{R}_{L,D}(f)\right|\\ 
&\quad +3\displaystyle\sup_{\|f\|_{H^{J_{2}}} \leq \lambda^{-1/2}} \left|\mathcal{R}_{L,P}(f)-\mathcal{R}_{L,D}(f)\right| +A_{2}^{J_{1}}(\lambda)+A_{2}^{J_{2}}(\lambda).
\end{align*}}%
First note that for $f \in \lambda^{-1/2}B_{H^{J}}$ and $B:=c_{L}(\lambda^{-1/2})\lambda^{-1/2}+1$, we have $|L(x,y,f(x))|\leq |L(x,y,f(x))-L(x,y,0)|+L(x,y,0)\leq B$ for all $(x,y) \in \mathcal{X}\times\mathcal{Y}$. Also note that the entropy bound assumption implies that $E_{D_{\mathcal{X}}\sim P^{n}_{\mathcal{X}}}\left(e_{i}(\lambda^{-1/2}B_{H},\|\cdot\|_{L_{\infty}(D_{\mathcal{X}})})\right)\leq \lambda^{-1/2}ai^{-\frac{1}{2p}}$.

Now note from Lemma \ref{lem:handhj} in section \ref{sec:featel} that the conditions of Lemma \ref{lem:rd-rp} in section \ref{sec:theory} are satisfied for $\mathcal{F}:=\lambda^{-1/2}B_{H^{J}}$, $\|\cdot\|_{\mathcal{F}}:=\|\cdot\|_{H^{J}}$, $C:= \lambda^{-1/2}$ and $B:=c_{L}(\lambda^{-1/2})\lambda^{-1/2}+1$ for each of the RKHS classes $H^{J}$. Also since $a^{2p}\leq \lambda^{p}n$ and $B\geq 1$, we have $\left(\frac{a^{2p}}{\lambda^{p}n}\right)^{1/2}\geq \left(\frac{a^{2p}}{\lambda^{p}n}\right)^{1/(p+1)}$ and $B^{1-p}\geq B^{\frac{1-p}{1+p}}$ for $p\in (0,1)$. Hence we have our assertion.
\end{proof}

\subsection{Proof of Lemma~\ref{lem:aims2}}\label{sec:aims2_proof}
\begin{proof}
\eqref{itm:aim1_svm} Fixing a $\lambda\in[0,1]$, we have that $B:=c_{L}(\lambda^{-1/2})\lambda^{-1/2}+1\leq 2\lambda^{-1/2}$. Now since $|X|\leq x \Rightarrow X\leq x$ for any $x\geq 0$, we see from Proposition \ref{prop:svm} that
{\small\begin{align}\label{eq:prop_res}
\mathcal{R}\mathcal{R}^D_D\left(J_1,J_2\right)&< A_{2}^{J_{1}}(\lambda) + A_{2}^{J_{2}}(\lambda) + 24\lambda^{-1/2}\sqrt{\frac{2\tau}{n}}+40\lambda^{-1/2}\frac{\tau}{n}+48K_{2}\lambda^{-\frac{p-1}{2}}\left(\frac{a^{2p}}{\lambda^{p}n}\right)^{\frac{1}{2}}\nonumber\\
&= A_{2}^{J_{1}}(\lambda) + A_{2}^{J_{2}}(\lambda) + 24\sqrt{2\tau}(\lambda n)^{-\frac{1}{2}} + 40\tau(\lambda^{\frac{1}{2}}n)^{-1}+48K_{2}a^{2p}(\lambda n)^{-\frac{1}{2}}
\end{align}}%
with probability at least $1-2e^{-\tau}$. Also from Corollary \ref{cor:svm} in section \ref{sec:theory}, for $J\in\widetilde{\mathcal{J}}$ similarly, we have
{\small\begin{align}\label{eq:cor_res}
&\left|\mathcal{R}\mathcal{R}^D_D\left(J\right)\right|< A_{2}^{J}(\lambda) + 12\sqrt{2\tau}(\lambda n)^{-\frac{1}{2}} + 20\tau(\lambda^{\frac{1}{2}}n)^{-1}+24K_{2}a^{2p}(\lambda n)^{-\frac{1}{2}}
\end{align}}
with probability at least $1-e^{-\tau}$. Now since $\lambda_{n} \rightarrow 0$ and $\displaystyle\lim_{n\rightarrow\infty}\lambda_{n}n=\infty$, Lemma $5.15$ along with $(5.32)$ of \citetalias{SVR} gives us that the right hand side of the above inequality converges to $0$. So the denseness assumption of the RKHSs additionally gives us the universal consistency of our feature elimination algorithm. To establish the convergence rate of our algorithm we further assume that there exists $c>0$ and $\beta \in (0,1]$ such that $A_{2}^{J}\leq c\lambda^{\beta}$ for any $J$ and for all $\lambda \geq 0$. Then it can be seen that asymptotically the best choice for $\lambda_{n}$ in \eqref{eq:prop_res} or \eqref{eq:cor_res} is a sequence that behaves like $n^{-\frac{1}{(2\beta+1)}}$ and then the inequalities in \eqref{eq:prop_res} and \eqref{eq:cor_res} are satisfied with the $l.h.s.$ replaced by $\epsilon_{n}$ and $\epsilon_{n}/2$ respectively, where $\epsilon_{n}$ is given by $(2c+24\sqrt{2\tau}+48K_{2}a^{2p})n^{-\frac{\beta}{2\beta+1}} + 40\tau n^{-\frac{4\beta+1}{2(2\beta+1)}}$. This proves \eqref{itm:aim1_svm} for $\{\epsilon_{n}\}$ for $\tau = o\left(n^{\frac{2\beta}{2\beta+1}}\right)$.

\eqref{itm:aim2_svm}
Observe from Corollary \ref{cor:svm} in section \ref{sec:theory} along with the conditions on $\lambda_{n}$, $A_{2}^{J}$, and steps in the proof of \eqref{itm:aim1_svm} given above that,
$ \left|\mathcal{R}\mathcal{R}^D_D\left(J\right)\right| < \epsilon_{n}/2$ occurs with $P^{n}$ probability greater than $1-e^{\tau}$ for any $J \subset \{1,2,\dots,d\}$ where $\epsilon_{n}$ is given as before.

Also note that from Assumption \ref{as:a2} in section \ref{sec:disc_assump} we have that $\mathcal{R}^{\ast}_{L,P,H^{J_{2}}} - \epsilon_{0}$ $\geq$ $\mathcal{R}^{\ast}_{L,P,H^{J_{\ast}}}$ $=$ $\mathcal{R}^{\ast}_{L,P,H^{J_{1}}}$. So for $H^{J_{2}}$ we have,
{\small\begin{align*}
&P^{n}\left(\left|\mathcal{R}\mathcal{R}^D_D\left(J_2\right)\right| < \epsilon_{n}/2\right) > 1 - e^{-\tau}\nonumber\\
\Rightarrow & P^{n}\left(\mathcal{R}^{\text{reg},\lambda_n}_{L,D,H^{J_2}}\left(f_{D,\lambda_{n},H^{J_{2}}}\right) + \epsilon_{n}/2 > \mathcal{R}_{L,P,H^{J_{2}}}^{\ast}\right) > 1 - e^{-\tau},
\end{align*}}
and for $H^{J_{1}}$ we have
{\small\begin{align*}
&P^{n}\left(\left|\mathcal{R}\mathcal{R}^D_D\left(J_1\right)\right| < \epsilon_{n}/2\right) > 1 - e^{-\tau}\nonumber\\
\Rightarrow &P^{n}\left(\mathcal{R}^{\text{reg},\lambda_n}_{L,D,H^{J_1}}\left(f_{D,\lambda_{n},H^{J_{1}}}\right) < \mathcal{R}_{L,P,H^{J_{1}}}^{\ast}+ \epsilon_{n}/2\right) > 1 - e^{-\tau}\nonumber\\
\Rightarrow &P^{n}\left(\mathcal{R}^{\text{reg},\lambda_n}_{L,D,H^{J_1}}\left(f_{D,\lambda_{n},H^{J_{1}}}\right) + \epsilon_{0} - \epsilon_{n}/2 < \mathcal{R}_{L,P,H^{J_{2}}}^{\ast}\right) > 1 - e^{-\tau}.
\end{align*}}
The above two statements then jointly imply that
{\small\begin{align}\label{eq:hj1-hj2}
\mathcal{R}\mathcal{R}^D_D\left(J_1,J_2\right) > \epsilon_{0}-\epsilon_{n}
\end{align}}
with $P^{n}$ probability greater than $1-2e^{-\tau}$.

Also it is easy to see that since $\epsilon_{n} \rightarrow 0$ with $n \rightarrow \infty$ for $\tau = o\left(n^{\frac{2\beta}{2\beta+1}}\right)$, the gap $\widetilde{\epsilon}_{n}=\epsilon_{0}-\epsilon_{n}$ $\longrightarrow$ $\epsilon_{0} >0$.

\eqref{itm:aim3_svm}
From Assumption \ref{as:a1} in section \ref{sec:disc_assump} and Corollary \ref{cor:oracle_svm} in section \ref{sec:theory}, conditions on $\lambda_{n}$, $A_{2}^{J}$, and steps in the proof of \eqref{itm:aim1_svm} given above, the `if' condition of \eqref{itm:aim3_svm} follows since for any $J$ and for all $\epsilon>0$, $\tau>0$ and $n\geq1$ we have,
{\small\begin{align}\label{eq:oracle_eq_svm}
P^{n}\left(\left|\mathcal{R}\mathcal{R}^D_D\left(J\right)\right| < \eta_{n}\right) > 1 - e^{-\tau},
\end{align}}
where $\eta_{n}=(c+8\sqrt{2\tau}+16K_{2}a^{2p})n^{-\frac{\beta}{2\beta+1}} + 40/3\tau n^{-\frac{4\beta+1}{2(2\beta+1)}}$.

Now for $J_{1}\in\widetilde{\mathcal{J}}$ and $J_{2}\notin\widetilde{\mathcal{J}}$ we have $\mathcal{R}^{\ast}_{L,P,H^{J_{2}}} - \epsilon_{0}$ $\geq$ $\mathcal{R}^{\ast}_{L,P,H^{J_{\ast}}}$ $=$ $\mathcal{R}^{\ast}_{L,P,H^{J_{1}}}$ $=\mathcal{R}^{\ast}_{L,P,H}$ and hence the `only if' condition of \eqref{itm:aim3_svm} also follows by noting that
$\lambda_{n}\left
\|f_{D,\lambda_{n},H^{J_{2}}}\right\|^{2}_{H^{J_{2}}} + \mathcal{R}_{L,P}\left(f_{D,\lambda_{n},H^{J_{2}}}\right) - \mathcal{R}_{L,P,H}^{\ast} > \epsilon_{0} - \eta_{n}$ occurs with $P^{n}$ probability greater than $1 - e^{-\tau}$.

Now since $\eta_{n} \rightarrow 0$ with $n \rightarrow \infty$ and $\tau = o\left(n^{\frac{2\beta}{2\beta+1}}\right)$, the gap $\widetilde{\widetilde{\epsilon}}_{n}=\epsilon_{0}-\eta_{n}$ $\longrightarrow$ $\epsilon_{0} >0$.
\end{proof}

\bibliography{rfe_ref}
\end{appendix}

\newpage

\begin{supplement}[id=sec:suppA]
\label{sup:supa}
  \sname{Additional Materials and Matlab Codes}

  \sdescription{Additional materials and details on the codes are given in the html page.}
	  \slink[url]{http://www.bios.unc.edu/~kosorok/RFE.html}
\end{supplement}

\renewcommand{\thesection}{S\arabic{section}}
\renewcommand{\thethm}{S\arabic{thm}}
\setcounter{section}{0}
\setcounter{thm}{0}

\section{A further discussion on Projected Spaces}\label{sec:appendkandk-j}

In order to provide a heuristic understanding of the importance of the projection spaces in feature selection, we give an alternative definition of lower dimensional versions of the input space. For a given input space $\mathcal{X}$, we can define the deleted space $\mathcal{X}^{-J}$ as the space spanned in $\mathbb{R}^{d-|J|}$, obtained by considering only the relevant $d-|J|$ co-ordinates of the vectors from $\mathcal{X}$. This is essentially different from the projected space $\mathcal{X}^{J}$, which amounted to looking at the span of the $d-|J|$ co-ordinates only when the remaining $|J|$ co-ordinates are 0. 

The original RFE procedure developed in \citet{Guyon2002} starts off with a given input space $\mathcal{X}$ and eliminates features by re-training the SVM algorithm on the lower dimensional spaces $\mathcal{X}^{-J}$. From their discussion, it is seen that if the Gram matrix of the training vectors $\{x_{1},\dots,x_{n}\}$ is given by $\displaystyle\{k(x_{k},x_{j})\}_{k,j=1}^{n}$, then the Gram matrix of the training vectors $\{x_{1}^{-i},\dots,x_{n}^{-i}\}$ after deleting a particular variable say $\mathcal{X}_{i}$ is taken to be $\displaystyle\{k(x^{-i}_{k},x^{-i}_{j})\}_{k,j=1}^{n}$. This clearly takes into account the assumption that the kernel $k$ can be defined on deleted vectors as well. It is not intuitively clear however if this can be done for any general kernel $k$. For example, if we start with $\mathcal{X} \subseteq \mathbb{R}^2$ and decide to look at only the first feature, then we are restricting ourselves to look at vectors only from $\mathbb{R}$. Suppose now that the original kernel $k$ was defined on $\mathbb{R}^2 \times \mathbb{R}^2$ as $k(x,y)=x^{\prime}Ay$ for a given $2\times 2$ matrix A. It is not clear then how it can be defined on $\mathbb{R} \times \mathbb{R}$ without breaking its form.  

However, for a radial or dot-product kernel, this connection can indeed be achieved. It can be shown that these special kernels have the ability to be redefined on deleted spaces quite easily, and in these cases, they become the same as their projected versions, that we propose to use. But for any general kernel, there might not be any ``natural" kernel to use in this regard. Hence, in this work we propose to use the projected space $\mathcal{X}^{J}$ instead of the deleted space $\mathcal{X}^{-J}$, as this approach allows a natural way to redefine the kernels in any general situation.

\section{Case Study 2 revisited: Naive rates for the Gaussian RBF kernel in the high dimensional scenario (\ref{itm:rc3} in Section \ref{sec:consis_d})} Now we look back at Case Study 2 to derive high dimensional rates for this problem. Being a universal kernel, the Gaussian RBF kernel produces a very rich RKHS. The entropy of such a space thus grow very slowly (at a much slower rate than the linear kernel) which makes learning so challenging. It is important to note that our algorithm learns at a rate that is directly dependent on the learning rate of the support vector machines algorithm itself, which is again directly dependent on the entropy growth of the RKHS (that we use for learning) with dimensionality. Our belief is that most of the bounds that we have on the complexity of an RKHS with the Gaussian RBF kernel in the literature, may have been derived under suboptimal conditions on the dimensionality growth. Thus deriving new bounds that are optimal in the dimensionality $d$ require new math which is certainly open for further exploration. For now, we concentrate on the bounds used in Case Study 2 in Section~\ref{sec:cs2}. \\

\textsc{The Entropy Bound:}
\begin{itemize}
\item We continue to use the entropy bound from Appendix \ref{sec:cs2_proof}. See that in the proof of Lemma \ref{lem:rbf_delta} we used $a:=c_{\epsilon,p}\gamma^{\frac{(1-p)(1+\epsilon)d}{2p}}$ for all $\gamma\geq 1$, for all $\epsilon>0$, $d/(d+\tau_d)<p<1$ and a constant $c_{\epsilon,p}$ depending only on $p$ and a given $\epsilon$, and where $\tau_d\in \left( \right. 0,\infty\left.\right]$ is the tail exponent of the distribution $P_\mathcal{X}$ on $\mathbb{R}^d$. 
\item Additionally assume there exist constant $K_1$ such that $\tau_d/d<K_1$. First note that this implies we can choose any $p \in [1/(1+K_1),1)$ and also that $(1-p)d<\tau_d d/(\tau_d+d)<d K_1/(1+K_1)$.
\item Note that $c_{\epsilon,p}$ is then a fixed constant, and we get $a<R_1 k_1^d$ for constants $R_1$ and $k_1$, or that $f(d)=k_1^d$.
\end{itemize}

\textsc{The Approximation Error Bound:}
\begin{itemize}
\item From Theorem $2.6$ of \citet{Steinwart07}, note that the geometric noise exponent $\alpha_{d_0}$ of the distribution $P$ on $\mathcal{X}\subset \mathbb{R}^{d_0}\times \mathcal{Y}$ can be expressed in terms of its Tsybakov noise exponent $q_{d_0}$ and its envelope $\gamma_{d_0}>0$ (refer to \citet{Steinwart07} for definitions of these quantities) as $\alpha_{d_0}=(q_{d_0}+1)\gamma_{d_0} d_0^{-1}$ for $q_{d_0}>1$.
\item We assume $q_{d_0}/d_0>C_1>0$ and $\gamma_{d_0}>\varepsilon>0$ for some constants $C_1$ and $\varepsilon$, which implies $\alpha_{d_0}>C_1 \varepsilon$. 
\item Note that we showed in Appendix \ref{sec:cs2_proof} that $A_{2}(\lambda,d_0,\gamma(\lambda,d_0))<K_{d_0}\lambda^{\beta}$ for $\beta=\alpha/(1+\alpha)$. Note that the earlier assumption then restricts $\beta$ to be strictly positive. 
\item The constant $K_{d_0}$ can be derived from \citet{Steinwart07} as $K_{d_0}=\max \left( \left(\frac{81}{\pi}\right)^{d/2}\theta(d)^{2}, 8 C (2d_0)^{\alpha d_0/2}\right)$ where $\theta(d)$ is the volume of the input space $\mathcal{X}$.
\item For $\mathcal{X}\subseteq r\mathbb{B}_1^{d_0}$ for some $r<\infty$ where $\mathbb{B}_1^{d_0}$ is the unit Euclidean ball in $\mathbb{R}^{d_0}$, we have $\theta(d)=O(r^{d_0})$.
\item Then for large enough $d_0$, $K_{d_0}<R_2 d_0^{k_2 d_0}$, and thus $g(d_0)=d_0^{k_2 d_0}$.
\end{itemize}

\section{Additional Proofs}

\subsection{Proof of Lemma~\ref{lem:linfty}}\label{sec:linfty_proof}
\begin{proof} The direction $\mathcal{L}_{\infty}^{J}(\mathcal{X}^{J}) \subseteq \mathcal{L}_{\infty}(\mathcal{X}^{J})$ is obvious since co-ordinate projection maps are continuous. To show that $\mathcal{L}_{\infty}^{J}(\mathcal{X}^{J}) \supseteq \mathcal{L}_{\infty}(\mathcal{X}^{J})$ let us take $g\in \mathcal{L}_{\infty}(\mathcal{X}^{J})$. Then $g\::\:\mathcal{X}^{J}\mapsto\mathbb{R}$ is measurable with $\|g\|_{\infty}<\infty$. Extend $g$ to $\widetilde{g}$ to include the whole domain $\mathcal{X}$ by defining $\widetilde{g}(x)=g\left(\pi^{J^{c}}(x)\right)$. Since $\widetilde{g}$ is measurable with $\|\widetilde{g}\|_{\infty}=\|g\|_{\infty}$, we have that $\widetilde{g} \in \mathcal{L}_{\infty}(\mathcal{X})$ and $\widetilde{g}\circ\pi^{J^{c}}=\widetilde{g}$, so $g=\widetilde{g}\big|_{\mathcal{X}^{J}} \in \mathcal{L}_{\infty}^{J}(\mathcal{X}^{J})$.
\end{proof}

\subsection{Lemma~\ref{lem:fandfj}}\label{sec:fandfj_proof}
Here we provide a lemma similar to Lemma \ref{lem:handhj} noting similar results for any general space. 
\begin{lem}\label{lem:fandfj} Let $\mathcal{F} \subset \mathcal{L}_{\infty}(\mathcal{X})$ be a non-empty functional subspace. Then for any $J \subseteq \{1,2,\dots,d\}$,
\begin{enumerate}
	\item If $\mathcal{F}$ is dense in $\mathcal{L}_{\infty}(\mathcal{X})$, then $\mathcal{F}^{J}$ is dense in $\mathcal{L}_{\infty}^{J}(\mathcal{X})$.\label{itm:dense_erm}
	\item If $\mathcal{F}$ is compact, then so is $\mathcal{F}^{J}$.\label{itm:comp_erm}
	\item $e_{i}(\mathcal{F}^{J},\|.\|_{\infty})$ $\leq$ $e_{i}(\mathcal{F},\|.\|_{\infty})$, $\forall i\geq 1$ where $e_{i}(\mathcal{F},\|.\|_{\infty})$ is the $i^{th}$ entropy number of the set $\mathcal{F}$ with respect to the $\|.\|_{\infty}$-norm as defined in Section~\ref{sec:consis}.\label{itm:covering_bound}
\end{enumerate}
\end{lem}
Condition \ref{itm:dense_erm} holds even if $\mathcal{L}_{\infty}(\mathcal{X})$ is replaced by any space that admits the nested structure. 
We provide a proof of the above lemma below, and the proof of Lemma \ref{lem:handhj} follows similarly.

\begin{proof}\eqref{itm:dense_erm} For any function $f$ $\in$ $\mathcal{L}_{\infty}(\mathcal{X})$, by the denseness of $\mathcal{F}$ we can find a sequence of functions $\{g_{n}\} \in \mathcal{F}$ such that $g_{n} \rightarrow f$ uniformly. Now fix an arbitrary function $f \in \mathcal{L}_{\infty}^{J}(\mathcal{X}) \subset \mathcal{L}_{\infty}(\mathcal{X})$ and consider any sequence of functions $\{g_{n}\}\in \mathcal{F}$ that converges to $f$ uniformly. Construct the new sequence of functions $\{g_{n}^{J}\}$ where for any function $f\in \mathcal F$, $f^{J}$ is defined by $f^{J}(x) = f(\pi^{J^{c}}(x))$. Observe trivially that $\{g_{n}^{J}\}$ $\in$ $\mathcal{F}^{J}$.\\ Now $\{g_{n}\} \mapsto f$ uniformly $\Rightarrow$ for any $\epsilon >0$, $\exists$ $N$ such that $\forall$ $n\geq N$,
\begin{eqnarray*}
&&\quad \displaystyle\sup_{x \in \mathcal{X}}|g_{n}(x) - f(x)| < \epsilon \quad \forall n \geq N\\
&\Rightarrow& \displaystyle\sup_{x \in \pi^{J^{c}}(\mathcal{X})}|g_{n}(x) - f(x)| < \epsilon \quad \forall n \geq N\\
&\Rightarrow& \quad \displaystyle\sup_{x \in \mathcal{X}}|g_{n}(\pi^{J^{c}}(x)) - f(\pi^{J^{c}}(x))| < \epsilon \quad \forall n \geq N\\
&\Rightarrow& \quad \displaystyle\sup_{x \in \mathcal{X}}|g_{n}^{J}(x) - f(x)| < \epsilon \quad \forall n \geq N \quad \quad (\because \quad f(\pi^{J^{c}}(x))=f(x)) \\
&\Rightarrow& \quad \{g_{n}^{J}\} \mapsto f \quad \text{uniformly}.
\end{eqnarray*}
Hence $\mathcal{F}^{J}$ is dense in $\mathcal{L}_{\infty}^{J}(\mathcal{X})$.

It is also easy to see that the above proposition will hold true even if $\mathcal{L}_{\infty}^{J}(\mathcal{X})$ is replaced by any other space $\mathcal{G}$ that admits the nested structure as long as we work with the uniform metric.

\eqref{itm:comp_erm} Since $\mathcal{F}$ is compact, for any $\epsilon$ $> 0$, $\exists$ $\{f_{n}\}_{n=1}^{N_{\epsilon}} \in \mathcal{F}$ such that $\mathcal{F}$ $\subset$ $\displaystyle \bigcup_{n=1}^{N_{\epsilon}}{\mathbb{B}_{\|\cdot\|_{\infty}}(f_{n},\epsilon)}$ (where $\mathbb{B}_{\|\cdot\|_{\infty}}(f_{n},\epsilon)$ is a $\|\cdot\|_{\infty}$ ball of radius $\epsilon$ with center $f_{n}$). We now fix $f$ $\in$ $\mathcal{F}^{J}$ and note that $\exists$ an equivalent class of functions $\{g^{f}\}$ in $\mathcal{F}$ such that for any two functions $g^{f}_{1}$ and $g^{f}_{2}$ $\in$ $\{g^{f}\}$ we have that $g^{f}_{1} \sim g^{f}_{2}$ in the sense that $g^{f}_{1}\circ\pi^{J^{c}} = g^{f}_{2}\circ\pi^{J^{c}} = f$. Fix one such $\widetilde{g}^{f}$ $\in$ $\{g^{f}\}$. Since $\widetilde{g}^{f} \in \mathcal{F}$, $\exists$ $f_{i}$ $\in$ $\{f_{n}\}_{n=1}^{N_{\epsilon}}$ such that $d(f_{i},\widetilde{g}^{f})<\epsilon$, that is,
\begin{eqnarray*}
&&\quad \displaystyle\sup_{x \in \mathcal{X}}|f_{i}(x) - \widetilde{g}^{f}(x)| < \epsilon\quad \Rightarrow \displaystyle\sup_{x \in \pi^{J^{c}}(\mathcal{X})}|f_{i}(x) - \widetilde{g}^{f}(x)| < \epsilon \\
&\Rightarrow& \quad \displaystyle\sup_{x \in \mathcal{X}}|f_{i}(\pi^{J^{c}}(x)) - \widetilde{g}^{f}(\pi^{J^{c}}(x))| < \epsilon\\
&\Rightarrow& \quad \displaystyle\sup_{x \in \mathcal{X}}|f_{i}^{J}(x) - f(x)| < \epsilon \quad (\because \: \widetilde{g}^{f}(\pi^{J^{c}}(x))=f(x)) \\
&\Rightarrow& \quad \{f_{n}^{J}\}_{n=1}^{N_{\epsilon}}\: \text{forms a finite} \: \epsilon \text{-cover for the set} \: \mathcal{F}^{J}.
\end{eqnarray*}
Hence $\mathcal{F}^{J}$ is compact.

\eqref{itm:covering_bound}
To see \eqref{itm:covering_bound}, note that if $f_{1}, \dots,f_{2^{n-1}}$ is an $\epsilon$-net of $\mathcal{F}$, then for any $f\in \mathcal{F}$, we have $i\in \{1,\dots,2^{n-1}\}$ such that $\|f-f_{i}\|_{\infty}<\epsilon$. Then,
\begin{eqnarray*}
\|f\circ\pi^{J^{c}}-f_{i}\circ\pi^{J^{c}}\|_{\infty} &=& \sup_{x \in\mathcal{X}}\left|f\circ\pi^{J^{c}}(x)-f_{i}\circ\pi^{J^{c}}(x)\right| = \sup_{x \in\mathcal{X}^{J}}\left|f(x)-f_{i}(x)\right|\\
&\leq& \sup_{x \in\mathcal{X}}\left|f(x)-f_{i}(x)\right|=\|f-f_{i}\|_{\infty}<\epsilon.
\end{eqnarray*}
Hence $f_{1}\circ\pi^{J^{c}}, \dots,f_{2^{n-1}}\circ\pi^{J^{c}}$ is an $\epsilon$-net of $\mathcal{F}^{J}$.
\end{proof}

\subsection{Proof of Lemma~\ref{lem:dot_nested}}\label{sec:dot_nested_proof}
\begin{proof}
To see this, let us consider a dot-product kernel $k$ such that $k(x,y)=g(\left\langle x,y\right\rangle)$ where $\left\langle \cdot\:,\:\cdot \right\rangle$ is the usual Euclidean inner-product. Now consider the pre-RKHSs $H_{\text{pre}}$ and $H^{J}_{\text{pre}}$.
We show here that $H^{J}_{\text{pre}}\subseteq H_{\text{pre}}$ which will imply that $H^{J}\subseteq H$. To show this, take $f \in H^{J}_{\text{pre}}$. This implies that $f$ can be written as $f(\cdot) = \displaystyle\sum_{i=1}^{n}\alpha_{i}k^{J}(\cdot,x_{i})$ for $n\in\mathbb{N},\:\alpha_{1},\dots,\alpha_{n}\in\mathbb{R},\:x_{1},\dots,x_{n}\in\mathcal{X}$. Hence,
\begin{eqnarray*}
f(\cdot) &=& \displaystyle\sum_{i=1}^{n}\alpha_{i}k^{J}(\cdot,x_{i}) = \displaystyle\sum_{i=1}^{n}\alpha_{i}k\left(\pi^{J^{c}}(\cdot),\pi^{J^{c}}(x_{i})\right) \\
&=& \displaystyle\sum_{i=1}^{n}\alpha_{i}g\left(\left\langle \pi^{J^{c}}(\cdot),\pi^{J^{c}}(x_{i})\right\rangle\right) = \displaystyle\sum_{i=1}^{n}\alpha_{i}g\left(\left\langle \cdot,\pi^{J^{c}}(x_{i})\right\rangle\right) \\
&=& \displaystyle\sum_{i=1}^{n}\alpha_{i}k\left(\cdot,\pi^{J^{c}}(x_{i})\right)
\end{eqnarray*}
Noting that $\pi^{J^{c}}(x_{1}),\dots,\pi^{J^{c}}(x_{n})$ $\in \mathcal{X}$, we have that $f \in H_{\text{pre}}$. In a similar way, we can show that for any $J_{1}\subseteq J_{2}$, $H^{J_{2}}\subseteq H^{J_{1}}$.
\end{proof}

\section{Real Data Examples} 

\begin{table}[h]
\scriptsize
\setlength{\tabcolsep}{2pt}
\centering
\begin{tabular}{l cc}
\hline\\
\multirow{2}{*}{\textbf{Method}}& Mean & No of\\
 & test error &  features chosen \\[2pt]
\hline\\
SVM wRisk-RFE & 0.08 & 4\\
SVM woSelection & 0.18 & 10\\
SVM wSCAD & 0.14 & 8\\
Log Reg wLASSO & 0.19 & 9\\[2pt]
\hline\\
\end{tabular}
\begin{center}	
\caption{Vowel Data%
	\label{tab:an_class}}
\end{center}
\end{table}

We apply the risk-RFE algorithm to two feature selection applications: feature selection in vowel recognition data, and feature selection in predicting ozone levels.

\subsection{Non linear classification in vowel recognition} To show the applicability of risk-RFE for non-linear classification, we apply our method to the vowel
recognition data set available at \url{http://www-stat.stanford.edu/~tibs/ElemStatLearn/} \citep[see][]{Hastie2009}. The dataset consists of recordings from $15$ individuals speaking different classes of vowels, $11$ in all, and a $10$ dimensional input space created from the reflection coefficients to calculate $10$ log area parameters \citep[see][]{Rabiner1978}. We apply SVM with a Gaussian RBF kernel to classify between the vowel sounds `i' and `I' based on these $10$ features. We use five fold cross-validation to obtain the SVM parameters and trained on a data set with $48$ instances and tested with $42$ instances of each vowel. We compared results of running risk-RFE with SVM Gaussian (SVM Gauss-wRFE), with (a) running SVM without any feature selection (SVM Gauss-woSelection), (b) running Logistic regression with LASSO and (c) SCAD wLinear SVM. The results are given in Table \ref{tab:an_class}. We did not use Guyon's RFE as it does not provide any inherent mechanism of choosing a subset of features. As can be seen, risk-RFE is able to achieve more than a $40\%$ drop in misclassification error relative to it's nearest competitor.

\begin{table}[t]
\scriptsize
\setlength{\tabcolsep}{2pt}
\centering
\begin{tabular}{l cccccccc}
\hline\\
\textbf{Variables} & Temp & invHt & hum & invTemp & press & milPress & vis & wind \\[2pt]
\hline\\
Value function & 39.90	& 35.18 &	30.37 &	29.16 &	28.56 &	28.19 &	27.89 & 27.42\\[2pt]
\hline\\
\end{tabular}
\begin{center}
\caption{SVM-wRFE: Ozone Data%
\label{tab:data_reg1}}
\end{center}
\end{table}


\subsection{Non linear regression in Ozone data}

The ozone data set \citep[see][]{Breiman1985} has been used frequently to evaluate methods for non-linear regression. These data record the level of atmospheric ozone concentration from eight daily meteorological measurements made in the Los Angeles basin in $1976$, and we have $330$ complete cases. The response, referred to as ozone, is actually the log of the daily maximum of the hourly-average ozone concentrations in Upland, California and can be found at \url{http://www-stat.stanford.edu/~tibs/ElemStatLearn/} \citep[see][]{Hastie2009}. We split the data randomly into twenty test and training sets of equal sizes and perform SVR with epsilon-insensitive loss and Gaussian RBF kernel on each training-test data pair. Five fold cross-validation was used to select the parameters of the algorithm in each run. The ranks of the variables as given by risk-RFE are provided in Table \ref{tab:data_reg1} with the respective objective function values in the last run. It does confirm important non-linear relationships existing between the predictors and the ozone levels which we already knew - that temperature ($Temp$) is the most important variable in predicting ozone levels followed by inversion height ($invHt$) and humidity ($hum$). Risk-RFE selects between $3$ and $4$ variables (either only these three, or these alongwith $InvTemp$) in each run, achieving an average test error rate of $15.38$\%. We also compared results from when we run the regression without using risk-RFE, and when we use linear regression with LASSO for feature selection, and the results are given in Table \ref{tab:data_reg}.

\begin{table}[h]
\scriptsize
\setlength{\tabcolsep}{2pt}
\centering
\begin{tabular}{l cc}
\hline\\
\multirow{2}{*}{\textbf{Method}}& Mean &Average no of\\
 & test error &  features chosen \\[2pt]
\hline\\
SVM wRisk-RFE & 15.38 & 3.6\\
SVM woSelection & 15.45 & 8\\
Lin Reg wLASSO & 19.74 & 5.8\\[2pt]
\hline\\
\end{tabular}
\begin{center}
\caption{Ozone Data %
	\label{tab:data_reg}}
\end{center}
\end{table}

\end{document}